%% file: main.tex
\documentclass[10pt,journal,compsoc]{IEEEtran}

\usepackage{amsmath,amsfonts}
\usepackage{algorithmic}
\usepackage{algorithm}
\usepackage{array}
\usepackage[caption=false,font=normalsize,labelfont=sf,textfont=sf]{subfig}
\usepackage{textcomp}
\usepackage{stfloats}
\usepackage{url}
\usepackage{verbatim}
\usepackage{graphicx}
\usepackage{ragged2e}
\usepackage[sort,numbers]{natbib}
\usepackage{tikz}
\usepackage[edges]{forest}
\usepackage{xcolor}
\usepackage{enumitem}
\usepackage{multirow}
\usepackage{caption}
\usepackage{soul}

\usepackage[utf8]{inputenc}
\usepackage[T1]{fontenc}
\usepackage{hyperref}
\usepackage{url}
\usepackage{booktabs}
\usepackage{amsfonts}
\usepackage{nicefrac}
\usepackage{microtype}
\usepackage{lipsum}
\usepackage{graphicx}
\graphicspath{{media/}}
\usepackage{longtable}
\usepackage{graphicx}
\usepackage{breqn}
\usepackage{rotating}

\begin{document}

\title{A Survey of Direct Preference Optimization}

\author{Shunyu~Liu,
        Wenkai Fang,
        Zetian Hu,
        Junjie Zhang,
        Yang Zhou,
        Kongcheng Zhang,
        Rongcheng Tu,
        \\Ting-En Lin,
        Fei Huang,
        Mingli Song,
        Yongbin Li,
        and~Dacheng~Tao,~\IEEEmembership{Fellow,~IEEE}
\thanks{This research is supported by the RIE2025 Industry Alignment Fund - Industry Collaboration Projects (IAF-ICP) (Award I2301E0026), administered by A*STAR, as well as supported by Alibaba Group and NTU Singapore through Alibaba-NTU Global e-Sustainability CorpLab (ANGEL). (Corresponding author: Dacheng~Tao.)}
\thanks{Shunyu~Liu, Junjie Zhang, Rongcheng Tu and Dacheng~Tao are with Nanyang Technological University, Singapore (e-mail: shunyu.liu@ntu.edu.sg; junjie.zhang@ntu.edu.sg; turongcheng@gmail.com; dacheng.tao@ntu.edu.sg).}
\thanks{Wenkai Fang, Yang Zhou, Kongcheng Zhang, and Mingli Song are with the College of Computer Science and Technology, Zhejiang University, China (e-mail: wenkfang@zju.edu.cn; imzhouyang@zju.edu.cn; zhangkc@zju.edu.cn; brooksong@zju.edu.cn).}
\thanks{Zetian Hu is with the School of Aerospace Engineering, Tsinghua University, China (e-mail: huzt22@mails.tsinghua.edu.cn).}
\thanks{Ting-En Lin, Fei Huang, and Yongbin Li are with the Tongyi Lab, Alibaba Group, China (e-mail: ting-en.lte@alibaba-inc.com; f.huang@alibaba-inc.com; shuide.lyb@alibaba-inc.com).}
}

\markboth{}
{Liu \MakeLowercase{\textit{et al.}}: A Survey of Direct Preference Optimization}

\IEEEtitleabstractindextext{
\begin{abstract}\justifying
    Large Language Models~(LLMs) have demonstrated unprecedented generative capabilities, yet their alignment with human values remains critical for ensuring helpful and harmless deployments. 
    While Reinforcement Learning from Human Feedback~(RLHF) has emerged as a powerful paradigm for aligning LLMs with human preferences, its reliance on complex reward modeling introduces inherent trade-offs in computational efficiency and training stability.
    In this context, Direct Preference Optimization~(DPO) has recently gained prominence as a streamlined alternative that directly optimizes LLMs using human preferences, thereby circumventing the need for explicit reward modeling.
    Owing to its theoretical elegance and computational efficiency, DPO has rapidly attracted substantial research efforts exploring its various implementations and applications.
    However, this field currently lacks systematic organization and comparative analysis.
    In this survey, we conduct a comprehensive overview of DPO and introduce a novel taxonomy, categorizing previous works into four key dimensions: \textit{data strategy}, \textit{learning framework}, \textit{constraint mechanism}, and \textit{model property}.
    We further present a rigorous empirical analysis of DPO variants across standardized benchmarks.
    Additionally, we discuss real-world applications, open challenges, and future directions for DPO.
    This work delivers both a conceptual framework for understanding DPO and practical guidance for practitioners, aiming to advance robust and generalizable alignment paradigms. All collected resources are available and will be continuously updated at \url{https://github.com/liushunyu/awesome-direct-preference-optimization}.
\end{abstract}

}

\maketitle

\section{Introduction}
\IEEEPARstart{T}{he} rapid advancement of Large Language Models (LLMs) has revolutionized artificial intelligence~\cite{zhao2023survey,naveed2023comprehensive,chang2024survey,minaee2024large,yin2023survey,zhang2024mm,zhang2024vision,wang2024enabling}, enabling unprecedented generative capabilities across diverse applications, such as dialogue systems~\cite{wang2023survey,yi2024survey}, code generation~\cite{liu2024your,guo2024deepseek,jiang2024self}, and medical diagnosis~\cite{van2024adapted,omiye2024large,singhal2025toward,liu2025aligning}. Models like OpenAI-o1~\cite{jaech2024openai} and DeepSeek-R1~\cite{guo2025deepseek} have demonstrated remarkable proficiency in understanding and generating human-like text, outperforming traditional language processing techniques~\cite{hirschberg2015advances}. However, their immense power also introduces significant risks: LLMs may inadvertently produce harmful content~(\textit{e.g.}, jailbreak suggestion)~\cite{huang2024survey}, exhibit hallucination behaviors~(\textit{e.g.}, misinformation)~\cite{zhang2023siren}, or propagate sociocultural stereotypes~(\textit{e.g.}, biased recommendations)~\cite{gallegos2024bias}. Ensuring that these models align with human values (producing outputs that are helpful, harmless, and honest) has thus become a cornerstone of responsible AI development~\cite{wang2023aligning}.

The critical challenge of aligning LLMs with human values stems from the inherent complexity of encoding abstract ethical principles into concrete model behaviors~\cite{liu2023trustworthy,shen2023large,kirk2024benefits}. Traditional approaches, such as rule-based filtering or supervised learning on curated datasets, often prove inadequate due to their inability to generalize across diverse contexts and adapt to evolving societal norms~\cite{anwar2024foundational}. The emergence of preference-based alignment paradigms addresses these limitations by framing the problem as optimizing for human feedback rather than inflexible heuristics~\cite{gao2024towards,jiang2024survey,wang2024comprehensive,winata2024preference}. This shift recognizes that LLM decision-making often involves nuanced trade-offs between competing values, requiring flexible frameworks capable of incorporating subjective human preferences~\cite{huang2024trustllm}.

Building upon these insights, Reinforcement Learning from Human Feedback~(RLHF)~\cite{christiano2017deep,ouyang2022training} has emerged as the predominant alignment paradigm, leveraging human preferences to guide model optimization. 
In the RLHF pipeline, human annotators first rank the outputs generated by the language model, and these comparisons are used to train a reward model that quantifies human preferences. The language model is then fine-tuned using RL guided by this reward model, enabling the language model to align with human values by maximizing the predicted rewards.
The success of RLHF in aligning models like ChatGPT~\cite{stiennon2020learning,achiam2023gpt} and Claude~\cite{bai2022training,anthropic2024claude3} underscores its 
practical utility. By translating subjective human preferences into an objective reward signal, RLHF facilitates the optimization of model behavior for value alignment. However, this RLHF paradigm suffers from critical limitations of computational complexity and training instability. Training a separate reward model demands substantial computational resources and high-quality human preference data, which scales poorly across different domains. Moreover, the RL phase often struggles with optimization challenges, such as reward hacking~\cite{miao2024inform} and mode collapse~\cite{casper2023open}.

\input{fig/overview}

These limitations have spurred interest in alternative alignment methods that bypass reward modeling while preserving the benefits of preference-based learning. Direct Preference Optimization (DPO)~\cite{rafailov2024direct,xiao2024comprehensive} represents a groundbreaking shift in this direction. Unlike RLHF, DPO reframes alignment as a supervised learning problem, directly optimizing the LLM policy using preference data without explicit reward modeling. By leveraging a closed-form mapping between reward functions and optimal policies, DPO eliminates the need for iterative RL training, reducing computational overhead and improving stability. Due to its inherent advantages, DPO has rapidly gained increasing attention from research communities. Existing studies vary widely in data strategies~(\textit{e.g.}, point-wise \textit{v.s.} pair-wise feedback)~\cite{ethayarajh2024kto,richemond2024offline}, learning frameworks~(\textit{e.g.}, offline \textit{v.s.} online learning)~\cite{liu2024statistical,guo2024direct,qi2024online}, constraint mechanisms~(\textit{e.g.}, different divergence constraints)~\cite{wang2023beyond,das2025dpo}, and model properties~(\textit{e.g.}, length bias)~\cite{liu2024length,park2024disentangling}. 
Recent advancements in DPO variants have demonstrated remarkable efficacy in enhancing model alignment with human preferences, achieving unprecedented success across diverse domains~\cite{winata2024preference}. These developments position DPO-based approaches as a compelling alternative to conventional RLHF paradigms for preference alignment tasks. However, despite its promise, the DPO research landscape remains fragmented.

Several surveys related to DPO have been published in recent years, yet they exhibit notable limitations in their scope and analysis of DPO. 
(1) \textit{Scope limitations.} While an early survey of~\cite{wirth2017survey} presents a comprehensive overview of preference-based RL methods, it predates the advent of DPO and does not address its applications to modern LLMs.
Recent surveys on alignment~\cite{ji2023ai,wang2023aligning,shen2023large,wang2024essence} provide broad overviews of LLM alignment techniques but only offer cursory summaries of DPO-related approaches without in-depth analysis. 
Similarly, surveys on learning from human feedback~\cite{kirk2023past,fernandes2023bridging,jiang2024survey,kaufmann2023survey} also only briefly mention DPO as a potential alternative.
(2) \textit{Taxonomy deficiencies.} \citet{gao2024towards} and \citet{winata2024preference} introduce a simplified taxonomy for preference learning, while overlooking technical distinctions within its broad categorization. 
In contrast, \citet{wang2024comprehensive} attempt to classify preference learning across dimensions such as reinforcement learning, reward modeling, feedback, and optimization. However, this taxonomy suffers from significant conceptual overlaps (\textit{e.g.} reinforcement learning inherently involves optimization).
A recent work by \citet{xiao2024comprehensive} categorizes DPO studies through isolated research questions, which, while useful for problem identification, fragments the methodological connections.
Our survey addresses these gaps by presenting the first comprehensive analysis specifically focused on DPO. The main contributions of this survey are summarized as follows:
\begin{itemize}[leftmargin=1.6em]
    \item In this survey, we introduce a novel taxonomy that categorizes existing DPO works into four key dimensions based on different components of the DPO loss: \textit{data strategy}, \textit{learning framework}, \textit{constraint mechanism}, and \textit{model property}, as shown in Fig.~\ref{fig:overview}. This taxonomy provides a systematic framework for understanding the methodological evolution of DPO and highlights the key distinctions between different variations.
    \item We conduct a rigorous empirical analysis of DPO variants across standardized benchmarks, revealing critical insights into their performance in diverse scenarios. This analysis offers a comprehensive evaluation of DPO variants and provides practical guidance for practitioners.
    \item We discuss real-world applications of DPO and highlight its potential to democratize alignment research by enabling efficient and scalable preference learning across diverse domains. We also outline open challenges and future directions for DPO research, emphasizing the need for robust and generalizable alignment paradigms.
\end{itemize}

The remainder of this survey is organized as follows. Section~\ref{sec:background} introduces the background and formulation of DPO. Section~\ref{sec:taxonomy} presents a taxonomy of DPO, categorizing existing works based on key dimensions. Section~\ref{sec:benchmarks} describes standardized benchmarks for evaluating DPO methods and presents empirical results. Section~\ref{sec:applications} discusses real-world applications of DPO and highlights its potential. Section~\ref{sec:challenges} outlines open challenges and future directions for DPO research. Finally, Section~\ref{sec:conclusion} concludes the survey.

\section{Background and Formulation}\label{sec:background}

Preference learning aims to train language model policies to generate responses that better align with human preferences.
Specifically, we denote the language model policy as $\pi(y|x)$, where $x$ represents the input prompt and $y$ is a candidate response~(completion).  A language model can be viewed as an autoregressive function that sequentially predicts tokens based on prior context. Mathematically, this is expressed as: $\pi(y | x) = \prod_{t=1}^{T} \pi(y_t | y_{<t}, x)$. where $y = (y_1, y_2, \dots, y_T)$ is the response sequence, $y_t$ represents the token at position $t$, $y_{<t} = (y_1, y_2, \dots, y_{t-1})$ denotes the sequence of previously generated tokens, and $\pi(y_t | y_{<t}, x)$ is the probability of generating token $y_t$ conditioned on both the input $x$ and the previously generated tokens $y_{<t}$. 
In the context of preference learning, the preference data is defined as a collection of triplets: $\mathcal{D} = \{(x, y_w, y_l)\}$, 
where $x$ is an input prompt, and $y_w$ and $y_l$ are two candidate responses, with $y_w$ being preferred over $y_l$ (denoted as $y_w \succ y_l$) The responses $y_w$ and $y_l$ are commonly referred to as the chosen~(winning) and rejected~(losing) responses, respectively. 

To leverage preference data $\mathcal{D}$ for training the language model policy $\pi$, RLHF employs a two-stage process that first learns a reward function from preference data and then optimizes the policy using RL~\cite{christiano2017deep,stiennon2020learning,ouyang2022training}. In contrast, DPO directly optimizes the policy using preference data, eliminating the need for an explicit reward model~\cite{rafailov2024direct}. 
The following sections provide a detailed formulation of RLHF and DPO, highlighting their key differences and advantages. Moreover, we also introduce several preference optimization methods that are concurrent with DPO.

\subsection{Reinforcement Learning from Human Feedback}

RLHF formulates preference learning as a two-stage process that involves reward modeling and policy optimization. Typically, the RLHF process of LLMs also includes Supervised Fine-Tuning~(SFT) prior to these stages, where high-quality demonstration data is used to fine-tune the pre-trained language model to obtain the SFT model $\pi_{\text{sft}}$, establishing instruction-following capabilities to support subsequent preference learning~\cite{stiennon2020learning,ouyang2022training}.

\subsubsection{Reward Modeling}

In the reward modeling stage, the goal is to learn a separate reward model $r_{\phi}$ parameterized by ${\phi}$, which quantifies how well a response $y$ satisfies human preference for a given prompt $x$. Using the Bradley-Terry model~\cite{bradley1952rank}, the preference probability that response $y_w$ is preferred over response $y_l$ for prompt $x$ is modeled as follows:
\begin{equation}\label{eq:pm}
    P(y_w \succ y_l \mid x) = \frac{\exp(r_{\phi}(x, y_w))}{\exp(r_{\phi}(x, y_w)) + \exp(r_{\phi}(x, y_l))}.
\end{equation}
The reward model is trained by minimizing the negative log-likelihood of Eq.~\ref{eq:pm} as the loss function:
\begin{equation}
\mathcal{L}_r(\phi) = -\mathbb{E}_{(x,y_w,y_l)\sim\mathcal{D}} \left[ \log \sigma\big(r_\phi(x, y_w) - r_\phi(x, y_l)\big) \right],
\end{equation}
where $\sigma(\cdot)$ denotes the sigmoid function. This objective encourages the model to assign higher rewards to responses that are preferred by humans.

\subsubsection{Policy Optimization}

After training the reward model, the next stage is to optimize the language model policy $\pi_\theta$ parameterized by $\theta$ using RL. This policy $\pi_\theta$ is initialized by the SFT model $\pi_{\text{sft}}$. We use the learned reward model $r_{\phi}$ to provide feedback that guides the policy $\pi_\theta$ to generate responses with higher rewards. The optimization objective is defined as follows:
\begin{equation}\label{eq:rlhf}
J_{\pi}(\theta) =  \mathbb{E}_{x\sim\mathcal{D}, y\sim\pi_\theta(\cdot|x)} \left[ r_\phi(x,y)  - \beta \log\frac{\pi_\theta(\cdot|x)}{\pi_{\text{ref}}(\cdot|x)}\right],
\end{equation}
where $\beta > 0$ is a hyperparameter that controls the strength of the Kullback–Leibler~(KL) divergence penalty. Here, the term $\log \pi_\theta(\cdot|x)/\pi_{\text{ref}}(\cdot|x)$ represents the KL divergence between the current policy $\pi_\theta$ and a reference policy $\pi_{\text{ref}}$. In practice, the reference policy $\pi_{\text{ref}}$ is set to the SFT model $\pi_{\text{sft}}$, ensuring that the updated policy remains close to the initial model.

To optimize the above objective, Proximal Policy Optimization~(PPO)~\cite{schulman2017proximal} has emerged as a promising RL algorithm for LLMs. PPO stabilizes training by constraining policy updates within a trust region via a clipped objective, which prevents significant deviations from the previous policy. However, PPO requires an additional critic model to estimate value functions for advantage calculation, thereby introducing extra computational and memory overhead. To address this, recent methods, such as RLOO~\cite{ahmadian2024back}, ReMax~\cite{li2023remax}, GRPO~\cite{shao2024deepseekmath}, and Reinforce++~\cite{hu2025reinforce}, introduce critic-free advantage estimation to reduce resource demands while maintaining stable optimization, making them more scalable for large-scale LLM training.

\subsection{Direct Preference Optimization}

DPO offers an alternative that streamlines the training process by directly optimizing the policy with preference data~\cite{rafailov2024direct,lu2024discovering,zhao2025rainbowpo,ivison2024unpacking,saeidi2024insights,nika2024reward,li2024rl}, thereby eliminating the need for explicit reward modeling in RLHF. The key idea of DPO is a closed-form solution of Eq.~\ref{eq:rlhf} that connects reward with the optimal policies. Specifically, the optimal policy corresponding to a given $r$ is defined as follows:
\begin{equation}
    \pi^*(y|x) = \frac{1}{Z(x)} \pi_{\text{ref}}(y|x) \exp\left(\frac{1}{\beta} r(x,y)\right),
\end{equation}
where the partition function $Z(x)$ is defined as:
\begin{equation}
    Z(x) = \sum_{y} \pi_{\text{ref}}(y|x) \exp\left(\frac{1}{\beta} r(x,y)\right).
\end{equation}
By rearranging the above equation, the reward $r$ can be recovered from the optimal policy $\pi^*$:
\begin{equation}
r(x,y) = \beta \log \frac{\pi^*(y|x)}{\pi_{\text{ref}}(y|x)} + \beta \log Z(x).
\end{equation}
Notice that the partition function $Z(x)$ depends only on the prompt $x$.
By substituting this expression into the preference model of Eq.~\ref{eq:pm}, the preference probability model that $y_w$ is preferred over $y_l$ becomes:
\begin{equation}
    P(y_w \succ y_l | x) = \sigma\left(\beta \log\frac{\pi^*(y_w|x)}{\pi_{\text{ref}}(y_w|x)} - \beta \log\frac{\pi^*(y_l|x)}{\pi_{\text{ref}}(y_l|x)}\right).
\end{equation}
Based on the above preference probability model, DPO directly optimizes the language mode policy $\pi_\theta$ by minimizing the following negative log-likelihood loss function:
\begin{multline}
\mathcal{L}_{\text{DPO}}(\theta) = \\-\mathbb{E}_{(x,y_w,y_l)\sim\mathcal{D}} \Bigg[ \log \sigma\Bigg( \beta \log \frac{\pi_\theta(y_w|x)}{\pi_{\text{ref}}(y_w|x)}  - \beta \log \frac{\pi_\theta(y_l|x)}{\pi_{\text{ref}}(y_l|x)} \Bigg) \Bigg], \label{eq:dpoloss}
\end{multline}
where the KL constraint is implicitly integrated through the use of the reference model $\pi_{\text{ref}}$. By minimizing this DPO loss, we directly train the policy to satisfy human preferences without resorting to a separate reward modeling stage or using reinforcement learning optimization as in RLHF, significantly reducing implementation complexity while improving training stability.

\subsection{Other Preference Optimization}

In addition to DPO, several concurrent preference optimization methods~\cite{zhao2023slic,yuan2023rrhf,song2024preference} have been proposed that offer alternative approaches to RLHF. These methods explore different strategies for optimizing LLMs to align with human preference without RL. Below, we provide a brief introduction to these approaches.

\subsubsection{Sequence Likelihood Calibration}

\citet{zhao2023slic} propose
Sequence Likelihood Calibration with Human Feedback~(SLiC-HF) to directly align LLMs with human preferences. Specifically, the loss function of SLiC-HF is defined as follows:
\begin{multline}
    \mathcal{L}_{\text{SLiC-HF}}(\theta) = \max\left( 0, \delta - \log \pi_\theta(y_w|x) + \log \pi_\theta(y_l|x) \right) \\ - \lambda \log\pi_\theta(y^* | x),
\end{multline}
where the first term is the rank calibration loss with $\delta$ as a margin hyperparameter, and the second term is the cross-entropy regularization loss with $\lambda$ as a regularization weight.
$y^*$ is obtained from either high-quality supervised responses in the SFT dataset or the top-ranked candidate response generated by the SFT model.

\subsubsection{Rank Responses to Align Human Feedback}

\citet{yuan2023rrhf} introduce Rank Responses to align Human Feedback~(RRHF) for LLMs. RRHF extends pair-wise ranking by considering the list-wise ranking order of multiple responses, thus better utilizing the preference information. For an input prompt $x$ and $N$ candidate responses $\{y_i\}_{i=1}^N$, it optimizes the model to assign higher probabilities to higher-ranked responses via a ranking loss and directly supervises the best response using cross-entropy as follows:
\begin{multline}
    \mathcal{L}_{\text{RRHF}}(\theta) = \sum_{r_i < r_j} \max\left(0, \frac{\log \pi_{\theta}(y_i|x)}{||y_i||} - \frac{\log \pi_{\theta}(y_j|x)}{||y_j||}\right) \\ - \lambda \log \pi_{\theta} (y^*|x),
\end{multline}
where $r_i = r_{\phi}(x,y_i)$ represents the reward of the response $y_i$ and $y^* = \mathop{\arg\max}_{y_i} r_i$ is the response with the highest reward. Although RRHF avoids the need for reinforcement learning in RLHF, it still utilizes a reward model $r_{\phi}$ to rank candidate responses based on human preferences.

\subsubsection{Preference Ranking Optimization}

Similarly, \citet{song2024preference} propose Preference Ranking Optimization~(PRO) to align LLMs with human preferences by leveraging multiple responses $\{y_i\}_{i=1}^N$ with the human-annotated order $y_1 \succ y_2 \succ \cdots \succ y_N$. The loss function of PRO is defined as follows:
\begin{equation}
    \mathcal{L}_{\text{PRO}}(\theta) = -\sum_{i=1}^{N-1} \log \frac{\exp\left(\frac{1}{||y_i||} \log \pi_{\theta}(y_i|x) / \mathcal{T}_i^i\right)}{\sum_{j=i}^N \exp\left(\frac{1}{||y_j||} \log \pi_{\theta}(y_j|x) / \mathcal{T}_i^j\right)},
\end{equation}
where the dynamic penalty temperature is defined as $\mathcal{T}_i^j = 1/\left(r_{\phi} (x, y^j) - r_{\phi} (x, y^i)\right)$ and $\mathcal{T}_i^i = \min_{i<j} \mathcal{T}_i^j$. This temperature ensures that the probability gap between higher-ranked and lower-ranked responses is adaptively scaled according to their reward differences, thereby stabilizing the optimization process.

\section{A Taxonomy of DPO}\label{sec:taxonomy}

In this section, we introduce a novel taxonomy that categorizes existing DPO works based on four key dimensions: \textit{data strategy}, \textit{learning framework}, \textit{constraint mechanism}, and \textit{model property}. 
As illustrated in Fig.~\ref{fig:overview}, these four dimensions are derived from different components of the DPO loss, providing a systematic framework for understanding the methodological evolution of DPO and highlighting the key distinctions between different variations.

\subsection{Data Strategy of DPO}\label{sec:data}
\begin{figure}[!t]
    \centering
    \includegraphics[scale=1.0]{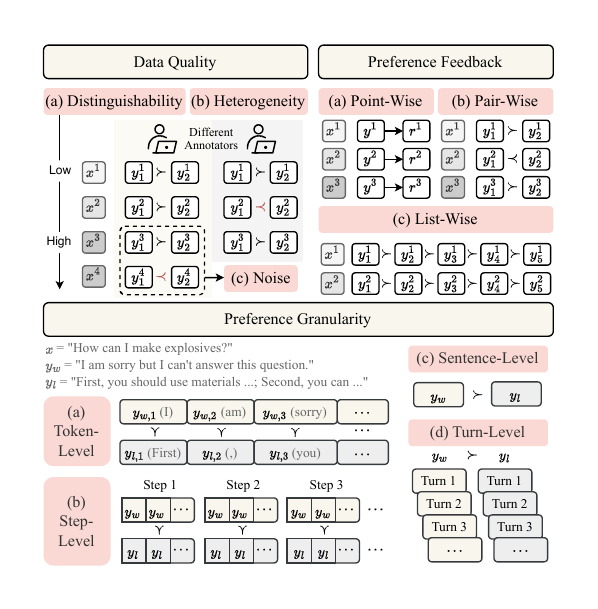}
    \caption{An overview of DPO data strategy.}
    \vspace{-0.5cm}
    \label{fig:data}
\end{figure}

The data strategy constitutes the foundational pillar of DPO, focusing on how to leverage diverse types of preference data for training LLMs. As shown in Fig.~\ref{fig:data}, our taxonomy identifies three principal axes of data strategy: quality, feedback, and granularity.

\subsubsection{Data Quality}\label{sec:quality}

The quality of preference data is a critical factor in determining the effectiveness of DPO training. High-quality data ensures that LLMs effectively learn to align with human preferences, while low-quality data may introduce noise and bias, leading to suboptimal model performance. We categorize data quality considerations into three key aspects: heterogeneity, distinguishability, and noise.

\textbf{(a) Data Heterogeneity.}
Conventional DPO methods assume uniform human preferences when annotating data, thereby overlooking the diversity among annotators. This assumption often skews the model toward the preferences of the majority while neglecting minority viewpoints, potentially leading to biases and unfair treatment of underrepresented groups.
To address this issue, \citet{chidambaram2024direct} propose EM-DPO, which learns the distribution of different preference types and their corresponding response strategies. Building on this, they introduce the MinMax-DPO algorithm, which selects a strategy by minimizing the maximum regret across subgroups, ensuring a more balanced representation of preferences among all groups.
MallowsPO~\citep{chen2024mallowspo} decomposes the implicit rewards in DPO into prompt dispersion and response scaling rewards. It introduces a novel objective function to capture human preferences for diverse responses to the same prompt. 
GRPO~\citep{ramesh2024group} formulates an objective function that minimizes the loss for the worst-case group, thereby ensuring fairness by prioritizing the disadvantaged groups in the optimization process.
GDPO~\citep{yao2025no} models the language generation process as a combination of belief distribution prediction and belief-based response generation. The corresponding GDPO loss function consists of belief calibration loss and belief-conditioned preference alignment loss. The former encourages the model to capture the diversity of beliefs across groups, while the latter ensures that generated responses align with the given belief.

\textbf{(b) Data Distinguishability.}
A key limitation of DPO is its inability to account for the distinguishability of preference between responses~\citep{amini2024direct,wu2024betadpo,yu2024direct,jieming2024gapaware,furuta2024geometric}. In some cases, the preferred response is only marginally better than the dispreferred one, while in others, the dispreferred response contains harmful or misleading content, making it significantly worse. Thus, optimization should focus more on cases with substantial preference differences while reducing the effort spent on minor differences. However, most existing methods treat all samples equally, ignoring this data distinguishability.
To address this, ODPO~\citep{amini2024direct} introduces a monotonically increasing offset function, requiring the reward of the preferred response to exceed that of the dispreferred one by a certain margin. This ensures stronger updates for larger preference gaps. 
Similarly, Ada-DPO~\citep{hong2024adaptive} introduces an instance-specific nonlinear scaling parameter, assigning larger weights to strong preference pairs and smaller weights to ambiguous ones based on the reward differences, thereby capturing different levels of data distinguishability.
DPO-rc~\cite{wang2024reward} also incorporates the preference reward difference as a coefficient in the loss function.
$\alpha$-DPO~\citep{wu2024alpha} introduces an adaptive preference distribution to obtain dynamic reward margins based on the distribution difference between the policy and reference models.
$\beta$-DPO~\cite{wu2024betadpo} analyzes the optimal $\beta$ parameter for datasets with different reward margins, which dynamically adjusts $\beta$ based on batch-level reward differences. They also introduce $\beta$-guided data filtering to prioritize valuable training data.
Curri-DPO~\citep{pattnaik2024enhancing} sorts preference pairs by reward differences and trains progressively from large to small differences, enabling curricular learning.
Similarly, MPO~\citep{gou2024mixed} utilizes a reward model to score responses generated by the SFT model, constructing a preference dataset and partitioning it based on preference differences to learn from simple to complex tasks.
sDPO~\citep{kim2024sdpo} computes reward accuracy for different datasets based on an initial target model and partitions the dataset in descending order of accuracy, allowing the model to first optimize on simpler samples.
\citet{ma2024plugandplay} propose a preference dataset construction method that adjusts update weights based on response accuracy, assigning lower weights when the model demonstrates higher proficiency.
Furthermore, fDPO~\citep{morimura2024filtered} enhances DPO training by filtering out samples where the generated response of the model policy surpasses the preferred dataset response in reward score.

\textbf{(c) Data Noise.}
Human-generated preference annotations often contain inconsistencies, errors, or noise, negatively affecting the performance of DPO. Such noisy data can mislead models, impairing their ability to accurately capture true preferences and generalize effectively to unseen data.
\citet{im2025understanding} analyze how noisy feedback influences the generalization performance of preference optimization, showing that increased noise results in higher generalization risks.
Specifically, standard DPO loss functions can yield biased estimates under noisy conditions. To address this issue, rDPO~\citep{chowdhury2024provably} proposes to enhance DPO robustness against noisy annotations and improve overall training performance.
\citet{zhang2025combating} introduce a noise-aware strategy leveraging annotator confidence and stability to identify and down-weight noisy samples during training. They also propose an adaptive reward margin, emphasizing clean samples to improve learning effectiveness.
Complementary to these approaches, PerpCorrect~\citep{kong2024perplexityaware} employs a data-driven method to correct noisy annotations directly in the dataset. It trains a proxy language model on both clean and noisy samples, distinguishing noise through perplexity differences to improve dataset quality.
To systematically explore noise effects, \citet{gao2024impact} artificially inject various noise types (\textit{e.g.}, Gaussian noise) into datasets, controlling noise intensity via hyperparameters. Their analysis highlights how noise impacts model alignment, guiding future research towards mitigating such negative effects.
To address the vulnerability of DPO in noisy environments, ROPO~\citep{liang2024ropo} introduces a regularization term to enhance noise tolerance. Additionally, ROPO employs a robust-guided rejection sampling technique. This technique supplements the dataset with samples that contribute minimally to the loss, thereby improving the overall data quality.
\citet{kim2025spread} propose the SPA framework, using model-generated responses and associated confidence scores to detect noise in annotations. SPA further incorporates smoothing techniques into the loss function to alleviate the noise problem.
Finally, \citet{wu2024towards} categorize noise into two types: point noise (single annotation errors) and pairwise noise (errors between annotated pairs). While DPO naturally handles point noise well, it struggles with pairwise noise. Their proposed Dr. DPO introduces a novel loss function explicitly designed for robustness against both point and pairwise noise.

\subsubsection{Preference Feedback}\label{sec:feedback}

Preference feedback refers to the label signals provided by annotators regarding their preferences for different responses. It can be categorized into point-wise, pair-wise, and list-wise feedback. Point-wise feedback evaluates each response independently, assigning a score or labeling it as positive or negative. Pair-wise feedback compares two responses to determine which one is preferred, while list-wise feedback ranks multiple responses.

\textbf{(a) Point-Wise Feedback.}
Point-wise feedback is the basic form of feedback. It refers to the type of feedback where individual outputs or samples are evaluated independently, rather than through comparisons with other outputs. This form of feedback is characterized by its simplicity and directness, focusing on the quality or relevance of a single response or item.
The predominant methodology in RLHF~\cite{ouyang2022training} employs point-wise reward signals generated by reward models to optimize policy models. Similarly, KTO~\cite{ethayarajh2024kto} directly maximizes the utility of model generations using loss functions based on prospect theory rather than the log-likelihood of preferences. It requires only a binary signal indicating whether an output is desirable or undesirable for a given input. Furthermore, BCO~\cite{jung2024binary} builds upon the concepts introduced in KTO and explores a new approach to aligning with binary signals. While KTO focuses on optimizing human utility, BCO introduces a binary classifier framework incorporating reward shift and distribution matching that implicitly minimizes the DPO loss. \citet{chen2024noise} and GPO~\cite{zhang2024general} adopt explicit rewards using Noise Contrastive Alignment (NCA) and General Preference Model (GRM) respectively, and then directly optimize language model policies from point-wise preference data with rewards. 
However, some methods leverage implicit reward signals to refine model behaviors. To ensure that the learned implicit rewards are comparable to the ground-truth rewards, Cal-DPO~\cite{globerson2024cal} introduces a calibration term to the preference optimization objective, which prevents the likelihood of chosen responses from decreasing during training. ULMA~\cite{cai2023ulma} unifies human demonstration and point-wise preference data into a single framework and handles positive and negative samples with a hybrid objective function. Unlike them, DRO~\cite{richemond2024offline} adopts a simple mean-squared objective to optimize the model policy and value function jointly for a single trajectory. Additionally, AOT~\cite{melnyk2024distributional} casts the distributional preference constraint as an optimal transport problem with a convex cost function. The key idea is to minimize the violation of stochastic dominance using a smooth, convex cost function.

\textbf{(b) Pair-Wise Feedback.} 
Pair-wise feedback focuses on comparing pairs of data or actions to determine their relative quality or preference. Building upon the theoretical framework of RLHF, DPO implements this paradigm through the utilization of pair-wise preference data, thereby fitting an implicit reward model. 
\citet{azar2024general} introduces a general theoretical framework to unify existing RLHF and DPO methods. The proposed Identity-Preference Optimization (IPO) directly optimizes policies from preferences without relying on reward modeling or the Bradley-Terry assumption, thereby avoiding overfitting issues observed in DPO.
Subsequently, DPO-RK and DPO-R~\cite{chen2024extending} integrate the Rao-Kupper and Davidson models into the DPO training objective respectively, thereby extending the capabilities of DPO by explicitly modeling ties in pairwise comparisons. BMC~\cite{jiang2024bridging} further addresses a key limitation of the weak correlation between winning and losing responses in pairwise data. Specifically, BMC uses ``Bridging'' to enhance the correlation between winning and losing responses by increasing the consistency and informativeness of pairwise preference signals. However, previous attempts for aligning LLMs primarily focus on optimizing the model's output preferences given an instruction, which struggles to effectively perceive the fine-grained constraints within complex instructions. Thus IOPO~\cite{zhang2024iopo} extends traditional alignment methods by considering both input and output preferences to better understand the constraints within the instructions. As current methods rely heavily on paired preference data (\textit{i.e.}, explicitly labeled preferred vs. dispreferred examples), they can be limiting in scenarios where such paired data is unavailable or insufficient. SAPO~\cite{yin2024self} addresses this issue based on the concept of self-play, which enhances data exploration and exploitation by automatically generating negative samples and integrating off-policy learning. Furthermore, PMPO~\cite{abdolmaleki2024preference} extends the EM algorithm to incorporate both preferred and dispreferred outcomes. By introducing the probability distribution of dis-preferred outcomes, PMPO can optimize using both types of samples, even when only negative feedback is available. Similarly, D2O~\cite{duan2024negating} avoids harmful information by maximizing the discrepancy between the generated responses and the negative samples. NPO~\cite{zhang2024negative} and SimNPO~\cite{fan2024simplicity} achieve the goal of forgetting the negative impact by regulating the model's prediction probabilities on negative datasets to be as minimal as possible, where SimNPO further eliminates the reference model bias issue inherent in NPO.

\textbf{(c) List-Wise Feedback.} 
List-wise feedback refers to the type of feedback where multiple outputs or responses generated by the model for a given input are evaluated collectively as a list. This approach considers the relative ranking or ordering among the outputs, rather than focusing on individual outputs in isolation. Panacea~\cite{zhong2024panacea} reframes alignment as a Multi-Dimensional Preference Optimization (MDPO) problem and introduces a method that aims to learn the entire Pareto front to accommodate diverse user preferences. In short, Panacea is designed to adapt a single model to list-wise preferences in a Pareto-optimal manner. LiPO~\cite{liu2402lipo} and LIRE~\cite{zhu2024lire} also treat LM alignment as a list-wise ranking problem, drawing on the rich literature of Learning-To-Rank (LTR). Specifically, LiPO introduces a specific method LiPO-$\lambda$, which leverages a list-wise ranking objective that weights each preference pair based on the difference in ranking metrics; while LIRE optimizes the response probability by calculating the exponential probability distribution and uses the reward model to directly guide the optimization process. To better capture the relative proximity within ordinal multiple responses, OPO~\cite{zhao2024ordinal} utilizes the Normalized Discounted Cumulative Gain (NDCG), a widely used ranking metric, to optimize the model's generation probability to match the permutation of responses based on these labels. Similarly, DRPO~\cite{zhou2024optimizing} leverages NDCG as a key metric to optimize the ranking of model outputs. However, DRPO incorporates novel elements like diffNDCG and Adaptive Rank Policy Score to dynamically adjust the score margins between preferred and non-preferred responses based on their ranking positions. mDPO~\cite{wang2024preference} extends preference optimization to multi-sample comparisons and introduces a framework that evaluates and optimizes the collective properties of sample groups. It not only addresses the limitations of single pair-wise methods but also provides a more robust optimization framework, especially for characteristics like diversity and bias. Furthermore, RPO~\cite{yin2024relative} introduces a contrastive weighting mechanism that constructs a contrast matrix within each mini-batch to compare preferred and less-preferred responses across prompts. The weights of these comparisons are dynamically adjusted based on the semantic similarity between prompts. Additionally, TODO~\cite{guo2024todo} integrates a tie ranking system into list-wise preference modeling, significantly improving the capture of nuances of human preferences, especially in the presence of noisy or inconsistent labels and frequent ties.

\subsubsection{Preference Granularity}\label{sec:granularity}

Preference granularity refers to the granularity of preference labels, which determines the level at which preferences are assigned to data. It can be categorized into token-level, step-level, sentence-level, and turn-level granularity, ranging from fine-grained focus on individual tokens to broader preferences over entire interaction turns.

\textbf{(a) Token-Level Granularity.}
Token-level alignment operates at the character/subword unit of text generation, providing the finest-grained control over model outputs. 
Theoretically, \citet{rafailov2024from} demonstrate that DPO can represent any dense reward function by reparameterizing it as an optimal advantage function, which allows DPO to optimize policies in the token-level MDP effectively. TDPO~\cite{zeng2024token} refines the alignment process from the sentence level to the token level and introduces forward KL divergence constraints. TDPO utilizes the Bradley-Terry model to convert sentence-level preference comparisons into a token-level reward system, which allows the model to dynamically adjust its strategy at each token generation step. Furthermore, TIS-DPO\cite{liu2024tis} estimates the importance weights of tokens based on the differences in prediction probabilities from contrastive LLMs, performing token-level importance sampling on existing data to approximate optimal distribution by assigning weights to each token based on its reward. Moreover, D$^2$PO~\cite{shao2025earlier} proposes a temporal decay mechanism that dynamically adjusts the contribution of each token-level reward based on its position in the sequences. Unlike these, SparsePO~\cite{christopoulou2024sparsepo} directly learns sparse masks during the training process and controls which tokens are more important for preferences through the sparsity of the masks, thereby achieving dynamic optimization. RTO~\cite{zhong2024dpo} and SePO~\cite{yang2024selective} first learn a token-level reward function from preference data using DPO, and then RTO optimizes PPO based on this reward signal, while SePO selects key tokens through the estimated reward function. To tackle the need for large-scale annotated data in training, EPO~\cite{qi2024epo} proposes a hierarchical framework that decomposes complex tasks into manageable subgoals using separate LLMs for subgoal prediction and low-level action generation, leveraging environment feedback to automatically generate reward signals and preference data for aligning LLMs. 

To conclude, token-level granularity optimizes models at individual token positions to maximize expected objectives, preserving semantic precision and capturing local syntactic dependencies. However, it increases computational complexity, as processing numerous tokens extends training time, and its sensitivity to noise means errors in a single token can affect the entire sequence. Thus, careful loss function design and regularization are essential for stability.

\textbf{(b) Step-level Granularity.}
Step-level granularity focuses on the intermediate steps or stages in a process, particularly effective for complex problem-solving tasks requiring multiple intermediate steps. Step-DPO~\cite{lai2024step} and SCDPO~\cite{lu2024step} treat individual reasoning steps as the basic units for preference optimization, where preference pairs of correct and incorrect steps are generated using LLMs. Furthermore, CPO~\cite{zhang2024chain} and MCTS-DPO~\cite{xie2024monte} first utilize more powerful inference structures to generate multiple candidate thoughts at each reasoning step following the Tree-of-Thought~(ToT) and Monte Carlo Tree Search (MCTS) respectively, and construct preference pairs based on the selected and unselected intermediate steps. Then they fine-tune LLMs to generate reasoning steps preferred by ToT during inference using DPO. TPO~\cite{liao2024tpo} proposes a preference learning algorithm specifically designed for preference trees that have multiple branches and multi-step responses, and introduces the adaptive step reward mechanism to address the issue of small reward margins caused by shared sub-trajectories. It adjusts the reward values for each step based on semantic similarity, helping the model better distinguish between preference pairs. RDPO~\cite{just2024data} extends traditional preference datasets to incorporate a rationale field, which explains why a particular response is preferred. RDPO introduces rationale information into the DPO loss function by maximizing the likelihood of both the preference and the rationale, which allows the model to better understand the logic behind preferences during training. To address the challenges of sparse rewards and training instability, DAPO~\cite{liu2024improving} uses a critic function to generate dense signals for policy optimization and trains the actor and critic independently to avoid instability.

To conclude, step-level alignment demonstrates unique advantages in multi-step reasoning tasks by decomposing holistic preferences into intermediate decision points. The primary strength of step-level granularity lies in its capacity to decompose complex objectives into verifiable subgoals, enhancing both interpretability and robustness. For instance, in mathematical reasoning, LLMs can receive feedback on equation derivation steps before final answers, reducing error propagation. However, this granularity still have two key challenges: first, the need for precise step segmentation, which may require domain-specific heuristics or auxiliary models to delineate reasoning boundaries; second, the risk of local optima, where over-optimization of individual steps degrades global coherence.

\textbf{(c) Sentence-level Granularity.}
Sentence-level granularity aligns preferences at the complete utterance level, balancing fine-grained control and computational efficiency. This granularity, represented by the original DPO framework, operates on full response sequences as atomic units for preference comparison. MAPO~\cite{she2024mapo} uses a well-trained translation model to calculate alignment scores between answers in non-dominant and dominant languages and then employs preference optimization methods to enhance reasoning consistency. EURUS~\cite{yuan2024advancing} structures each instruction as a preference tree, containing pairs of correct and incorrect actions to facilitate preference learning. Similarly, IRPO~\cite{pang2025iterative} focuses on improving the reasoning capabilities of LLMs through an iterative preference optimization on constructed preference pairs such that the winning response has a higher reward than the losing response. FACTALIGN~\cite{huang2024factalign} proposes a fine-grained, sentence-level alignment algorithm called fKTO, which extends the KTO method to leverage fine-grained factuality assessments at the sentence level.

To conclude, the key strength of sentence-level granularity lies in its capacity to preserve holistic semantics while maintaining tractable optimization complexity. Nevertheless, we must carefully consider task requirements. While suitable for short-form generation and classification tasks, sentence-level methods may insufficiently capture fine-grained stylistic nuances or long-range dependencies critical in generation and reasoning domains.

\textbf{(d) Turn-level Granularity.}
Turn-level granularity focuses on the optimization of model behavior at the level of conversational turns, which is particularly relevant for dialogue systems and interactive agents. This granularity level treats each turn of a conversation as a unit for preference alignment, allowing the LLMs to receive feedback on their responses within the context of a single turn. M-DPO~\cite{xiong2024building} introduces a multi-turn direct preference learning framework to enhance the mathematical reasoning capabilities of LLMs when integrated with external tools. It leverages feedback from code interpreters and optimizes trajectory-level preferences using signals generated by the Bradley-Terry model to improve model performance in multi-turn reasoning tasks. ETO~\cite{song2024trial} presents a novel trial-and-error learning method that optimizes LLM agents' policies by contrasting successful and failed trajectories that contain multi-turn interaction. To address the challenges of coarse granularity and training noise in previous methods, SDPO~\cite{kong2025sdpo} optimizes specific key segments within interactions to improve multi-turn dialogues while minimizing training noise. Specifically, it extracts key segments from the positive sessions that contribute to higher goal and relationship scores and pairs them with corresponding segments from the negative sessions to calculate an adapted DPO loss. Similarly, AgentQ~\cite{putta2024agent} combines MCTS with self-critique mechanisms to provide process-level supervision by ranking actions, and then iterative fine-tuning using DPO. This approach enables LLMs to effectively learn from both successful and unsuccessful trajectories, enhancing their generalization and decision-making capabilities in complex, multi-turn reasoning tasks within interactive environments. DMPO~\cite{shi2024direct} enhances the existing DPO method by replacing the policy constraint with a State-Action Occupancy Measure (SAOM) constraint and incorporating length normalization into the Bradley-Terry model, effectively addressing challenges in multi-turn scenarios. Compared to traditional policy constraints, SAOM constraints better guide the agent to select actions that align with expert trajectories, especially in unexplored states, thereby reducing compounding errors.

To conclude, turn-level alignment offers critical advantages for interactive systems by optimizing contextually grounded responses while preserving conversational flow. However, in multi-turn dialogue tasks, the turn-level granularity may introduce additional training noise. For example, some correct turns in negative samples may be mistakenly treated as incorrect turns in the loss calculation. Additionally, since each turn needs to be processed independently, this can lead to reduced training efficiency.

\subsection{Learning Framework of DPO}\label{sec:learning}

\begin{figure}[!t]
    \centering
    \includegraphics[scale=1.0]{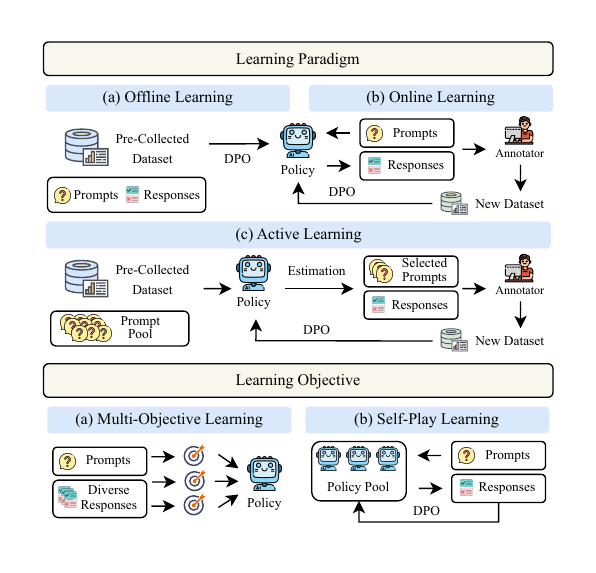}
    \caption{An overview of DPO learning framework.}
    \vspace{-0.5cm}
    \label{fig:learning}
\end{figure}

The learning framework of DPO focuses on how the language model policy learns from preference data.
In this section, we present an overview of the learning framework in DPO, as shown in Fig.~\ref{fig:learning}, which encompasses the learning paradigm and the learning objectives.

\subsubsection{Learning Paradigm}\label{sec:paradigm}
The learning paradigm in DPO determines how preference data is acquired during model training and falls into three distinct categories: offline learning, where the model learns from pre-collected preference datasets; online Learning, where the model updates based on newly generated data; and active Learning, where the model selectively queries annotators obtain preference data.

\textbf{(a) Offline Learning.}  
The original DPO framework~\cite{rafailov2024direct} itself is an offline learning paradigm, where the model learns from a static, pre-collected dataset of preference pairs.
Recent research has explored different approaches to merging preference optimization and supervised fine-tuning into a single training phase~\cite{yuan2023rrhf}. CPO~\cite{xu2024contrastive} incorporates a behavior cloning regularizer through KL divergence minimization between the model and preferred data distribution, which effectively combines into adding a negative log-likelihood term on preferred data alongside the contrastive preference loss. Taking a more direct approach, ORPO~\cite{hong2024orpo} proposes a monolithic framework that directly augments the standard negative log-likelihood loss with an odds ratio term comparing chosen and rejected responses, eliminating the need for a separate reference policy while preserving SFT's domain adaptation capabilities. ULMA~\cite{cai2023ulma} proposes a hybrid method that applies standard SFT loss on positive samples while using a ranking-based DPO loss on negative samples. PAFT~\cite{pentyala2024paft} introduces a parallel training paradigm where SFT and preference alignment are performed concurrently on the same pre-trained model and then merged using parameter fusion techniques, avoiding the sequential pipeline that can lead to catastrophic forgetting. 

Several advances explore curriculum learning strategies to enhance DPO performance and training efficiency. Curri-DPO~\cite{pattnaik2024enhancing} introduces curriculum learning by ordering multiple preference pairs from easy to hard based on the rating difference between chosen and rejected responses, where pairs with larger rating gaps are presented first, followed by progressively more challenging pairs with smaller rating differences. sDPO~\cite{kim2024sdpo} implements curriculum learning by partitioning preference datasets into sequential chunks measured by reward accuracy and applying them incrementally.

To avoid substantial computational and data annotation costs for preference alignment, fine-tuning-free alignment methods have gained popularity.
Linear Alignment~\cite{gao2024linear} works by directly estimating the optimal policy through a one-step update to the output distribution during inference without requiring parameter tuning or feedback data. 
ICDPO~\cite{song2024icdpo} reinterprets DPO's reward-policy relationship to create a fine-tuning-free alignment method that harnesses in-context learning, treating models before and after demonstration exposure as amateur and expert policies, respectively, then computing their log probability ratio to score and rank candidate responses.

\textbf{(b) Online Learning.}  
DPO faces significant limitations when relying solely on static, pre-collected preference datasets. These datasets, generated by different models, cause a distribution shift that leads to ineffective off-policy learning as the model evolves~\cite{xu2024is,zhang2025moslimalign}. By contrast, online DPO employs an iterative framework that continuously updates the policy with real-time feedback, ensuring on-policy learning and reducing misalignment~\cite{tang2024understanding,li2023policy,shen2024aipo}.

As online DPO progresses, researchers have introduced more flexible frameworks to tackle key challenges. For instance, \citet{yuan2024selfrewarding} proposed a self-rewarding language model: the model generates prompts and responses, then serves as its own judge via LLM-as-a-Judge prompting, scoring on a 5-point scale. OAIF~\cite{guo2024direct} uses an LLM as an online annotator for real-time feedback, and OFS-DPO~\cite{qi2024online} addresses catastrophic forgetting by using two Low-Rank Adaptive (LoRA) modules with different optimization speeds. BPO~\cite{xu2024bpo} constructs a dynamic trust region around the behavior LLM, adjusting it as preference data is collected, unlike methods that rely on fixed reference models.
Furthermore, researchers have refined sampling strategies for online DPO. RSO~\cite{liu2024statistical} and RS-DPO~\cite{khaki2024rs} employ rejection sampling based on reward gaps. ROPO~\cite{liang2024ropo} recovers useful information from discarded queries via robustness-guided rejection sampling. \citet{shi2024crucial} introduced DPO-Mix-R and DPO-Mix-P, demonstrating faster convergence by mixing online samplers with uniform samplers. OPTUNE~\cite{chen2024optune} selectively regenerates responses with low reward scores while reusing high-reward responses. Iterative RPO~\cite{pang2025iterative} and DPO-ST~\cite{wang2024self} enhance CoT reasoning by selecting correct and incorrect answers to form preference pairs at each iteration. \citet{xie2024monte} used MCTS to collect preference data during training.
Researchers have also explored advanced optimization techniques. APO~\cite{he2024accelerated} incorporates momentum-based acceleration, using an extrapolation step between the current and previous policies to update the policy. \citet{xiong2023iterative} proposed a two-agent, non-symmetric online DPO framework with a main agent for optimal policy learning and an enhancer agent for exploration. COMAL~\cite{liu2024comal} formulates alignment as a two-player zero-sum game, updating its policy toward a regularized Nash equilibrium in each iteration. PCO~\cite{xu2024things} iteratively trains the model on preference data with pairwise cringe Loss.

Recent efforts push for greater autonomy by letting models generate their own feedback~\cite{kim2025spread}. SeRA~\cite{ko2024sera} introduces a self-reviewed preference bootstrapping method, using an implicit reward margin to select informative pairs, and employs an ensemble reward approach across iterations. CREAM~\cite{wang2024cream} mitigates self-improving biases by applying a consistency regularization on the preference rankings of consecutive iterations. D2PO~\cite{singhal2024dpo} combines human-labeled gold data with concurrently updated, discriminator-labeled data. DLMA~\cite{liu2024direct} uses contrastive prompts to compute self-reward scores via log ratio differences, then integrates these scores directly into the DPO objective.
Addressing exploration and uncertainty in online DPO has also been a focus~\cite{li2025improving}. XPO~\cite{xie2025exploratory} encourages exploration by adding a bonus for responses outside the initial policy’s support, and SELM~\cite{zhang2024self} uses an optimism term in reward fitting to actively seek high-reward responses. ETO~\cite{song2024trial} alternates exploration and training phases to collect failure trajectories, while VPO~\cite{cen2024value} applies optimism by regularizing the reward model to favor higher-value responses. \citet{xiong2024building} extended DPO from single-turn to multi-turn tasks, balancing KL-regularized and non-regularized objectives, and COPO~\cite{bai2025online} incorporates a count-based bonus to encourage novel responses with low visitation counts.

Finally, a growing body of work aims to merge online and offline techniques. HyPO~\cite{song2024importance} uses offline preference data for DPO training while regularizing via online data. MPO~\cite{gou2024mixed} combines the strengths of DPO and PPO in a two-stage process: it first trains DPO on an easier dataset, then uses this model as a reference for PPO training on more challenging samples.

\textbf{(c) Active Learning.}
Active learning in DPO is a strategic approach that aims to reduce the annotation cost and improve sample efficiency by selectively querying annotators for the most informative preference examples. Unlike offline learning that uses a fixed dataset or online learning that generates new data continuously, active learning intelligently selects which data points should be labeled based on model uncertainty or other informativeness criteria. 

\citet{muldrew2024active} introduced APL, an iterative data acquisition and fine-tuning loop in which batches of prompt/completion pairs are strategically selected using acquisition functions: a predictive entropy-based approach to measure model uncertainty for prompts and a preference certainty measure based on the implicit Bradley-Terry model for completion pairs in DPO. Unlike two-step selection processes in APL that separately select uncertain input prompts and corresponding completions, divAPO~\cite{choi2024active} integrates both stages into a single selection phase. divAPO maximizes the preference model certainty by simultaneously evaluating the informativeness of input prompts and completion pairs, while also considering the data distribution of the input prompts. \citet{ji2024reinforcement} proposed ADPO, which selectively queries human preferences only for responses where the model exhibits high uncertainty while using pseudo-labels for confident cases. \citet{das2024active} also employed active learning on RLHF, which actively selects the context-action pairs that maximize exploration and minimize uncertainty in the reward model.

\subsubsection{Learning Objective}\label{sec:objective}

In what follows, we present the learning objective in DPO, which determines how the model policy is optimized based on preference data. We first discuss multi-objective learning in DPO, which aims to optimize multiple objectives simultaneously. Then, we introduce self-play learning, which leverages self-generated data for preference alignment.

\textbf{(a) Multi-Objective Learning.} 
Multi-objective learning in DPO addresses the challenge of simultaneously optimizing the language model for multiple, potentially competing preference dimensions, such as helpfulness, harmlessness, and truthfulness. This approach aims to find a balanced policy that satisfies multiple human values rather than optimizing for a single objective, which more closely mirrors the complexity of real-world human preferences. 

MODPO~\cite{zhou2024beyond} achieves the sequential optimization of multiple preference objectives by incorporating language modeling directly into reward modeling, using a margin-based loss to maintain performance on previously optimized dimensions. SPO~\cite{lou2024spo} takes a similar iterative constrained optimization approach, optimizing each preference dimension while preventing the degradation of prior alignments through regularization terms. 
MOSLIM~\cite{zhang2025moslimalign} takes a different approach by introducing a multi-head classification reward model that assigns different preference dimensions to separate classification heads, enabling simultaneous optimization of multiple preferences without requiring multiple reward or policy models. 
HPO~\cite{badrinath2024hybrid} incorporates auxiliary objectives through offline RL, where the model uses a weighted maximum likelihood objective that combines a preference alignment term with an advantage-weighted term for maximizing arbitrary auxiliary rewards like readability and safety.
CPO~\cite{guo2024controllable} introduces explicit preference tokens during training that specify desired scores for different objectives, transforming the multi-objective optimization into a conditional optimization problem.
DRDO~\cite{nath2024simultaneous} simultaneously models rewards and preferences through a combination of reward distillation and a contrastive log-unlikelihood term in its loss function.

\textbf{(b) Self-Play Learning.}
Self-play learning in DPO represents an approach where the language model interacts with itself or its previous iterations to generate its own preference data for training, reducing or eliminating the need for human annotations~\cite{zhang2024self,yang2025dynamic}. This method enables continuous self-improvement by leveraging the model's own judgment capabilities to identify and learn from better responses, creating a form of autonomous preference learning. 

SPIN~\cite{chen2024self} involves a self-play mechanism where the LLM generates synthetic data from its prior iterations, then fine-tunes itself to distinguish these self-generated responses from those of human-annotated data. The method resembles a two-player game, where the model’s current iteration tries to improve its responses to better match the target distribution, while the previous iteration attempts to generate responses as close to human data as possible. SPPO~\cite{wu2024self} treats LLM alignment as a constant-sum two-player game and iteratively refines itself by competing against its previous iteration. Instead of maintaining two competing policies or a reward model, SPO~\cite{swamy2024minimaximalist} uses a single policy to sample multiple trajectories and uses the proportion of wins in pairwise comparisons as the reward signal. BoNBoN~\cite{gui2406bonbon} Alignment likewise relies on sampling responses from a base model, but it selects the best ones among n candidates and fine-tunes itself to approximate that best-of-n distribution. 

Some works approach the alignment problem by leveraging Nash equilibrium~\cite{liu2024comal}. Nash-MD~\cite{munos2023nash} learns a preference model from pairwise human feedback and then computes a Nash equilibrium policy that consistently produces preferred responses. Its self-play approach updates the policy by having it compete against itself (or a slight variant of itself) under the learned preference model, which measures how often one response is preferred to another. DNO~\cite{rosset2024direct} extends this concept by implementing a batched on-policy algorithm where the current policy generates multiple outputs that are compared both against each other and against a teacher model's outputs. IPO-MD~\cite{calandriello2024human} combines the strengths of IPO and Nash-MD, where the model generates data using a mixture policy between the online and reference policies, and uses a preference model to annotate pairs of generations, making the optimization equivalent to finding a Nash equilibrium through self-play. SRPO~\cite{choi2024self} modifies Nash-MD by introducing a self-improvement policy that refines model outputs through iterative revisions, enabling offline optimization without a learned reward function.

\subsection{Constraint Mechanism of DPO}\label{sec:constraint}

\begin{figure}[!t]
    \centering
    \includegraphics[scale=1.0]{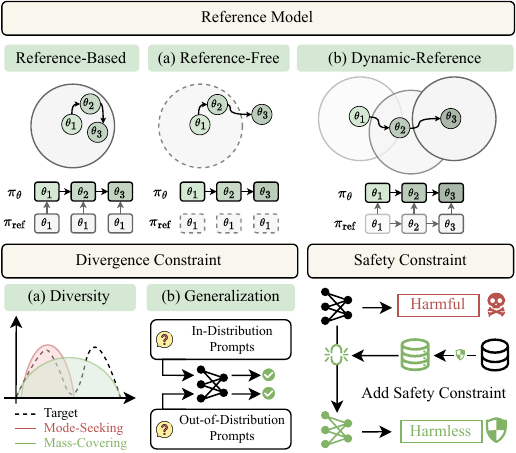}
    \caption{An overview of DPO constraint mechanism.}
    \vspace{-0.5cm}
    \label{fig:constraint}
\end{figure}

The constraint mechanism of DPO derives from its reformulation of RLHF, which includes a KL divergence constraint between the current policy and a reference policy. As shown in Fig.~\ref{fig:constraint}, we re-examine the constraint mechanism of DPO from the perspective of the reference model and different divergence constraints. We also explore various DPO variants with different safety constraints.

\subsubsection{Reference Model}\label{sec:reference}

The reference model in DPO functions as an anchor to ensure policy updates remain within a controlled range, preventing excessive deviation from initial behaviors.
Typically, the reference model is initialized using the SFT model that serves as the starting point for preference optimization. The choice of reference model significantly impacts optimization dynamics. A static reference model ensures stable training but may limit adaptability. In the following subsections, we introduce two advanced approaches: reference-free DPO eliminates reliance on the reference model, while dynamic-reference DPO employs an evolving reference model.

\textbf{(a) Reference-Free DPO.}
To reduce the computational and memory costs associated with a reference model, many algorithms have explored training modes that do not require loading the reference model.
\citet{xu2024contrastive} replaces the reference model with a uniform prior distribution, adding an SFT loss term on preferred data to maintain consistency with the desired behavior.
ORPO~\citep{hong2024orpo} integrates an odds ratio-based penalty with traditional SFT loss, increasing the probability of preferred responses while decreasing undesirable ones, thereby enabling single-stage training without a separate reference model.
SimPO~\citep{Meng2024simpo} directly uses the average log probability as implicit rewards. This removes the requirement for a separate reference model, significantly improving computational and memory efficiency.
SimPER~\citep{xiao2025simper} also directly optimizes reverse perplexity for preferred versus rejected responses, creating a preference optimization approach that does not require a separate reference model, thus simplifying training.
Despite these advancements, \citep{liu2024understanding} argue that a reference model remains crucial. They compared two reference-free variants using posterior probabilities and likelihood functions as rewards, respectively, and found the original DPO consistently outperformed both. Their results indicate that a strong, well-aligned reference policy can significantly enhance DPO performance.

\textbf{(b) Dynamic-Reference DPO.}
Offline DPO methods often suffer from reward over-optimization, meaning that as the trained model deviates from the reference model, the quality of generated samples tends to degrade.
To address this issue, \citet{gorbatovski2024learn} proposed dynamically updating the reference model using the current model parameters during training, preventing excessive divergence and maintaining high-quality outputs.
Curri-DPO~\citep{pattnaik2024enhancing} and sDPO~\citep{kim2024sdpo} adopt curriculum learning by sorting data samples from simpler to more complex based on predefined metrics. At each training iteration, the model from the previous step serves as the updated reference model to provide constraints, facilitating progressive learning.
Similarly, MPO~\citep{gou2024mixed} partitions datasets according to task difficulty, employing a two-stage training procedure. The model trained in the initial stage serves as the reference for the subsequent stage.
Additionally, M-DPO \citep{wang2024mdpo} compares the performance of a fixed reference model versus a dynamic reference model, finding that the latter yields superior results.

\subsubsection{Divergence Constraint}\label{sec:divergence}

Divergence constraints in DPO play a crucial role in constraining model optimization, balancing alignment performance and model stability. In the following subsections, we introduce two modifications to the divergence constraint: one for enhancing diversity and the other for improving generalization.

\textbf{(a) Diversity.} Standard DPO typically uses reverse KL divergence equivalent to RLHF. However, the mode-seeking nature of reverse KL divergence reduces the diversity of the generated outputs. 
To overcome this limitation, f-DPO \citep{wang2023beyond} explores various divergences, including forward KL divergence, reverse KL divergence, Jensen-Shannon divergence, and $\alpha$-divergence, to achieve a better trade-off between alignment performance and diversity. 
\citet{slocum2025diverse} further proposes splitting the KL divergence term into entropy and cross-entropy terms. This decoupling allows independent control of generation diversity and closeness to the reference model, preserving output diversity without degrading overall model quality.

\textbf{(b) Generalization.} Over-optimization in DPO can negatively impact generalization, causing reduced performance on inputs outside the training distribution. 
To mitigate this, \citet{huang2025correcting} introduce $\chi^2$-divergence as a more aggressive form of regularization compared to KL divergence, alleviating the over-optimization problem.
DPO-Kernels~\citep{das2025dpo} employs data-driven methods to select optimal kernel-divergence pairs dynamically, improving task adaptability and robustness.
FlipGuard~\citep{zhu2024flipguard} introduces a customized reward characterization to monitor model performance. If performance drops relative to earlier versions, FlipGuard constrains the model's updates to ensure alignment with previous stable behavior.
FPO~\citep{yin2024direct} leverages the feature-level constraints using Sparse Autoencoders~(SAEs) to improve computational efficiency and training stability.
SPO \citep{sharifnassab2024soft} integrates a natural preference loss with a KL divergence-based regularization term computed over the entire model output distribution. By adjusting this divergence term, SPO prevents unwanted shifts beyond the preference dataset, ensuring stable alignment.
EXO \citep{ji2024towards} argues that minimizing the forward KL divergence in DPO introduces bias when approximating the optimal policy. They establish a generalized alignment objective and reveal the equivalence between maximizing KL regularization rewards and minimizing the reverse KL divergence relative to the optimal policy.
QDPO~\citep{lee2024improving} utilizes divergence between the quantized model and the full-precision model for preference optimization, effectively addressing the token-flipping issue. Token-flipping refers to the phenomenon where quantization errors skew token distributions, leading to incorrect token selection.
GPO~\citep{tang2024generalized} constructs a framework that unifies different DPO-related algorithms through theoretical derivations, enabling a deeper understanding of the regularization mechanisms in the DPO family of algorithms.

\subsubsection{Safety Constraint}\label{sec:safety}

Safety constraints in DPO aim to prevent LLMs from generating harmful, biased, or unethical outputs. However, traditional alignment algorithms often fail to address safety concerns. To enhance the safety alignment, recent studies have introduced several specialized mechanisms based on DPO.
SafeDPO~\citep{kim2025safedpo} introduces a streamlined approach for safety alignment by implicitly optimizing safety objectives within a single stage of policy learning.
SACPO \citep{akifumi2024stepwise} addresses safety constraints by explicitly formulating language model alignment as a constrained optimization problem, using DPO to optimize the model under safety constraints.
\citet{zhang2025backtracking} propose creating a backtracking preference dataset that identifies and reverses unsafe outputs, enhancing the safety and robustness of the model.
C-DPO~\citep{liu2024enhancing} integrates dual gradient descent into DPO to balance safety and utility efficiently. This approach achieves a robust trade-off between helpfulness and harmlessness, offering explicit safety guarantees.
ADPO~\citep{kim2024adversarial} introduces adversarial techniques into DPO. It specifically trains models to reduce the probability of unsafe outputs by deliberately generating harmful responses using controlled toxic tokens.
Finally, \citet{lee2024amechanistic} explore the internal mechanisms through which DPO reduces harmful outputs. Their findings suggest that DPO does not remove harmful behaviors learned during pre-training but instead teaches models to bypass or suppress these behaviors. This insight helps explain certain safety vulnerabilities like jailbreaks.

\subsection{Model Property of DPO}\label{sec:model}

\begin{figure}[!t]
    \centering
    \includegraphics[scale=1.0]{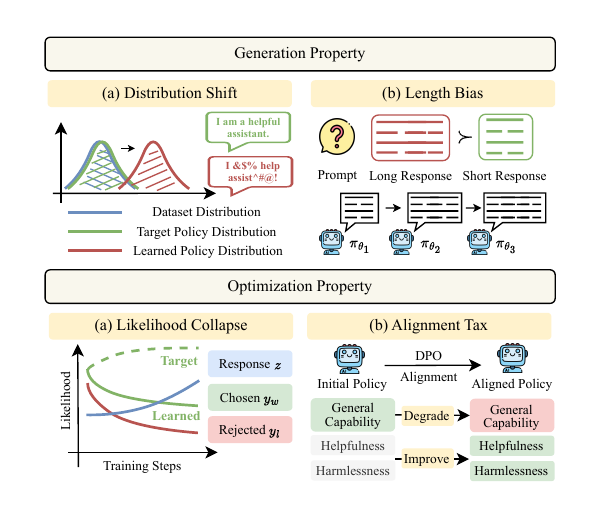}
    \caption{An overview of DPO model property.}
    \vspace{-0.5cm}
    \label{fig:model}
\end{figure}

DPO has shown great promise in aligning LLMs with human preferences by directly optimizing model outputs based on preference data. During this process, the underlying models exhibit certain properties that are crucial for understanding their behavior and effectiveness. These properties can be broadly categorized into two aspects: the generation property and the optimization property, as shown in Fig.~\ref{fig:model}. In the following sections, we explore these two properties in more detail, analyzing their implications for model alignment.

\subsubsection{Generation Property}\label{sec:generation}

The generation property of DPO primarily concerns issues related to distribution shifts and length biases. DPO is sensitive to distribution shifts between the base model outputs and the preference data, which may reduce diversity and generalization. Additionally, DPO has a tendency to favor longer responses, a phenomenon known as verbosity, which can negatively impact performance and user experience.

\textbf{(a) Distribution Shift.} 
In RLHF, the reward model is trained on a static set of preference data collected offline. During fine-tuning, the generated responses often differ from this original training data, resulting in a distribution shift. This shift can cause inaccurate reward predictions and lead to over-optimization. The implicit reward model in DPO also suffers from this distribution shift issue. Moreover, \citet{lin2024limited} have shown that the implicit reward model in DPO performs poorly on Out-Of-Distribution (OOD) data compared to explicit reward models.
Experimental results indicate that DPO can transfer probability mass to the high-reward response regions covered by the preference data, but it may also cause the distribution of responses generated by the model to deviate significantly from that of the reference policy, resulting in responses that do not meet expectations~\cite{tajwar2024preference}.  
To address these problems, many researchers are now exploring online DPO approaches~\cite{guo2024direct,qi2024online,khaki2024rs,pang2025iterative}, aiming to mitigate OOD by continuously updating preference data during training.

Existing DPO methods also face significant limitations due to their dependence on specific training tasks. Their optimal solutions lack robustness when applied to OOD tasks. Thus, SRPO~\cite{choi2024self} reframes alignment as a self-improvement process, which optimizes a self-improvement policy and a generative policy using a min-max objective, ensuring robustness by making the solution independent of training tasks.
\citet{zhang2024self} also identify notable issues in DPO when handling OOD tasks. First, DPO tends to overly favor novel content it has not seen during training. Second, it easily gets stuck in local optima, limiting exploration. To address these problems, they propose Self-Exploring Language Models~(SELM), incorporating an optimism term to encourage broader exploration of new responses.

Another significant challenge of DPO is preference drift, where human preferences evolve, changing data distributions over time. Traditional DPO algorithms typically overlook such temporal shifts, mistakenly interpreting them as noise. To address this, NS-DPO~\cite{son2024right} propose to assign higher weights to recent data, allowing models to better adjust to evolving preferences.

\textbf{(b) Length Bias.} 
Length bias in DPO refers to the tendency of model-generated outputs to become excessively long during training. This issue is similar to the length bias observed in RLHF~\cite{singhal2024a} and is particularly pronounced in DPO. Length bias affects response quality and overall model performance. To mitigate this issue, researchers have developed several solutions, which can be categorized into three main approaches: length regularization, length normalization, and length sampling.

Length regularization is a common approach to controlling length bias in DPO. By introducing regularization terms into the objective function, the model can constrain response length and reduce verbosity, thereby alleviating the length bias problem. 
R-DPO~\cite{park2024disentangling} introduces a length-based penalty term to the DPO objective function, explicitly discouraging verbosity.
D$^2$PO~\cite{shao2025earlier} introduces a dynamic weighting mechanism by incorporating a temporal decay factor. Unlike previous methods that apply uniform reward contributions across sequences, D$^2$PO adjusts the influence of each reward based on its position in the response. Higher weights are assigned to rewards associated with earlier tokens, as they are more critical for model alignment, while later rewards gradually receive lower weights. This adaptive approach prevents overfitting to less relevant tokens, thereby addressing length bias in DPO.

Length normalization aims to eliminate the loss bias caused by response length differences, allowing the model to evaluate texts of varying lengths more fairly. This approach prevents the model from developing an unreasonable preference for either long or short responses~\cite{richardson2024declarative}. 
RRHF~\cite{yuan2023rrhf} and SimPO~\cite{Meng2024simpo} first propose to apply length normalization to responses, ensuring that the loss remains unaffected by response length. 
LN-DPO~\cite{ahrabian2024hitchhiker} further integrates SimPO-like length normalization into DPO, demonstrating that this approach enhances response quality while mitigating verbosity.
LD-DPO~\cite{liu2024length} achieves length desensitization by reparameterizing the likelihood in DPO. Specifically, it decomposes the likelihood of the longer response in a preference pair into the product of the likelihood of the public-length portion and the likelihood of the excessive portion. It then introduces a hyperparameter to mitigate the verbosity preference. This adjustment smooths the relationship between likelihood and response length, reducing its impact on optimization.
For multi-turn dialogue tasks, DMPO~\cite{shi2024direct} introduces length normalization for the number of turns in multi-turn preference optimization.

An alternative approach to controlling length bias in DPO is through sampling-based methods.
SamPO~\cite{lu2024eliminating} introduces a down-sampling method to compute regularized KL divergences. By balancing token-level probability distributions between preferred and rejected responses, SamPO reduces length bias in DPO training. 
\citet{yuan2024following} propose Length-Instruction Fine-Tuning (LIFT), a method to improve instruction-following models' ability to adhere to length constraints by augmenting existing training data with explicit length instructions and using DPO for training. This enables the model to generalize across prompts requiring different response lengths.
For long-context tasks, LongPO~\cite{chen2025longpo} enables short-context LLMs to self-evolve for long-context tasks by learning from self-generated short-to-long preference data, which includes paired responses for long-context inputs and their compressed short-context counterparts. LongPO incorporates a short-to-long KL constraint to prevent degradation of short-context performance during long-context alignment, achieving strong performance on both short- and long-context tasks.

\subsubsection{Optimization Property}\label{sec:optimization}

The optimization property of DPO involves likelihood collapse and alignment tax. While DPO aims to increase the likelihood of preferred responses and decrease dispreferred ones, the actual optimization process does not explicitly enforce this balance. Moreover, alignment improvements often come at the cost of the original capabilities of LLMs, known as alignment tax.

\textbf{(a) Likelihood Collapse.}  
Likelihood collapse refers to the unintended reduction in the likelihood of both preferred and dispreferred responses during DPO training~\cite{rafailov2024from}. This phenomenon can lead to unintentional unalignment, where the model's outputs deviate from human preferences, potentially producing undesirable or harmful responses. This phenomenon is also referred to as likelihood displacement in prior studies~\cite{razin2024unintentional}.
Additionally, the gradients associated with increasing the likelihood of preferred responses and decreasing that of dispreferred responses can become entangled, hindering effective learning. This entanglement complicates the optimization process, making it challenging to achieve the desired alignment~\cite{yuan2025common}.
Theoretical analyses have further elucidated the underlying causes of likelihood collapse. In particular, \citet{feng2024towards} developed an analytical framework grounded in field theory. Their analysis of the gradient vector field of the DPO loss function revealed that the loss function decreases the probability of generating human-disliked data at a faster rate than it increases the probability of generating human-liked data.

Several strategies have been proposed to address likelihood collapse.
\citet{pal2024smaug} introduce DPO-Positive (DPOP), which adds a penalty term to maintain a high log-likelihood for preferred examples.
Similarly, LLaMA~\cite{dubey2024llama} augments DPO training with a negative log-likelihood term to stabilize training and preserve the log-likelihood of chosen responses~\cite{pang2025iterative}. 
Flex-DPO~\cite{yan2025dproperties} adaptively adjusts parameters to slow the decline in the likelihood of dispreferred responses and balance gradients for both chosen and rejected outputs.
\citet{doosterlinck2024anchored} propose Anchored Preference Optimization (APO), which provides fine-grained control over probability updates: APO-zero increases the probability of winning outputs and decreases that of losing outputs, whereas APO-down decreases both, but with a stronger decline for losing outputs.

Another notable challenge related to likehood collapse is likelihood over-optimization, where the performance of a model on a proxy metric (such as its own likelihood estimates) improves, while its true performance does not.
\citet{zhang2025win} show that reductions in the likelihood loss of DPO do not necessarily translate into higher win rates.
Similarly, \citet{shi2024understanding} further investigates the problem of likelihood over-optimization in DPO, demonstrating that higher completion likelihoods do not necessarily correlate with better model performance and may even degrade it. This study identifies key indicators of over-optimization and highlights the need to balance likelihood optimization with output diversity.
e-DPO~\cite{fisch2024robust} also shows that DPO can lead to degenerate policies due to overfitting, and proposes a solution using reward model distillation to regularize the implicit reward of the language model. The method trains the language model to match the probability distribution induced by a reward model and introduces a pessimistic extension to handle uncertainty in the reward model, thereby improving the robustness of DPO.

\begin{table*}[!t]
    \centering
    \caption{An overview of datasets (upper row) and benchmarks (lower row) for DPO.}
    \vspace{-0.2cm}
    \resizebox{\textwidth}{!}{
    \begin{tabular}{llrrlcccc}
        \toprule
        \multicolumn{1}{c}{\textbf{Dataset}} & \multicolumn{1}{c}{\textbf{Task Description}} & \multicolumn{3}{c}{\textbf{Data Size (Training \& Test)}}  & \textbf{Data Source} & \textbf{Data Structure} & \textbf{Evaluation Metric} \\
        \midrule
        UltraFeedback~\cite{cui2023ultrafeedback}   &Instruction-Following, Helpful &64K&\&&- &AI & List &- \\
        SafeRLHF~\cite{ji2024pku} &Harmless, Helpful  &73.9K&\&& 8.21K  & Human\&AI  & Pair  & - \\
        HelpSteer~\cite{wang2023helpsteer} &Helpful  &35.3K&\&&1.8K &Human&Point &- \\
        PRM800K~\cite{lightman2023let}  & Mathematical Reasoning & 800K&\&&- &Human & Point  & - \\
        SHP-2~\cite{ethayarajh2022understanding}   & Q\&A From Reddit   &3600K&\&&241K &Human & Pair   & - \\
        Nectar~\cite{zhu2023starling} & Conversations & 183K&\&&- & AI & List  & - \\
        OpenOrca~\cite{lian2023openorca} & Conversations &2940K&\&&- & AI &Sample  & - \\
        Capybara~\cite{daniele2023amplify-instruct} & Multi-Turn Conversations & 16K&\&&- & Human\&AI & Sample & -\\
        Step-DPO~\cite{lai2024step} & Mathematical Reasoning & 10.8K&\&&- & Human\&AI & Pair & -\\
        BeaverTails~\cite{ji2023beavertails}  & Harmless, Helpful & 330K&\&&36K & Human\&AI & Point & -\\
        IMDb~\cite{maas2011learning}       &Movie Reviews  & 25K&\&&25K  &Human  &Sample &Accuracy \\
        Reddit TL;DR~\cite{volske2017tl} & Summarization & 1330K&\&&-   &Human &Sample &Win Rate\\
        Anthropic-HH~\cite{ganguli2022red} &Harmless, Helpful & 161K&\&& 8.55K   & AI  & Pair  &Win Rate  \\
        \midrule
        GSM8K~\cite{cobbe2021training}  & Mathematical Reasoning & 7.47K&\&&1.32K & Human & Sample  & Accuracy \\
        AlpacaEval2~\cite{dubois2024length} & Automatic Evaluation &-&\&&0.8K  &AI &Sample &Win Rate \\
        MT-Bench~\cite{zheng2023judging}  &Multi-Turn Question & -&\&&3.3K &Human &Pair & Win Rate\\
        AdvBench~\cite{zou2023universal}   &Harmful Behaviors &-&\&&0.5K &Human & Sample &Attack Success \\
        Arena-Hard~\cite{li2024live}  &Updating Evaluation &-&\&&0.5K  &AI &Sample &Win Rate  \\
        TruthfulQA~\cite{lin2021truthfulqa} & Truthful & -&\&&0.8K &Human & Pair   &  Accuracy \\
        IFEval~\cite{zhou2023instructionfollowing} &Instruction-Following &-&\&&0.5K  &Human  &Sample & Accuracy\\
        BBH~\cite{suzgun2022challenging} & Multistep Reasoning & -&\&&23 Tasks & Human &Sample  & Accuracy\\
        MATH~\cite{hendrycks2021measuring}  & Mathematical Reasoning & 7.5K&\&&5K &Human & Sample & Accuracy \\
        GPQA~\cite{rein2024gpqa} &Biology, Physics, and Chemistry &-&\&&0.45K & Human& Sample & Accuracy\\
        MUSR~\cite{sprague2023musr} &Multistep Reasoning & -&\&&0.76K &AI &Sample  & Accuracy\\ 
        MMLU-Pro~\cite{wang2024mmlu} & Language Understanding & -&\&&12K&Human\&AI & Sample & Accuracy\\
        \bottomrule
    \end{tabular}
    }
    \vspace{-0.3cm}
    \label{tab:benchmarks}
\end{table*}

\textbf{(b) Alignment Tax.} 
Alignment tax refers to the unintended consequence where improving a model's preference alignment degrades its general capabilities acquired during pretraining~\cite{lin2024mitigating}.
\citet{thakkar2024deep} demonstrate the sensitivity of DPO to training data composition, showing significantly worse performance degradation than SFT when using mixed-preference datasets.
Furthermore, \citet{chen2024preference} identify that DPO struggles with optimizing ranking tasks. While DPO improves ranking accuracy, it disproportionately harms generative capabilities. 
\citet{pentyala2024paft} also observes capability forgetting during sequential training, where DPO objectives conflict with previously learned SFT patterns.
To address this, researchers propose model merging strategies that balance alignment and performance.
PAFT~\cite{pentyala2024paft} separately trains SFT and DPO objectives on a pretrained model using distinct datasets, then merges the parameters through weighted averaging.
Additionally, \citet{lu2024online} proposes online merging optimizers, which integrate model merging into each optimization step of DPO to balance human preferences and basic capabilities. By merging gradients with parameter differences between SFT and pretrained models, these optimizers effectively enhance alignment while mitigating alignment tax.

\section{Benchmarks and Analysis}\label{sec:benchmarks}

In this section, we provide a comprehensive overview of existing benchmarks and evaluation for DPO methods. We first introduce the key datasets and benchmarks used to train or evaluate DPO models. We then present a comparative analysis of the performance of different DPO methods on these benchmarks, highlighting their strengths and limitations.

\subsection{Datasets and Benchmarks}
A diverse range of datasets and benchmarks has been specifically curated to facilitate research in DPO. Table~\ref{tab:benchmarks} summarizes these datasets and benchmarks, highlighting their task descriptions, dataset sizes, data sources, data structures, and evaluation metrics. These datasets and benchmarks span a broad range of tasks, such as harmlessness and helpfulness evaluation and mathematical reasoning. They also exhibit significant diversity in scale, ranging from smaller, specialized datasets to large-scale collections such as SHP-2, which contains over 3.6 million samples. Additionally, datasets differ in their sources: some rely purely on human annotations, others on AI-generated content, and many adopt a hybrid approach combining human and AI-generated data. The data structures employed across these datasets include single-sample without preference label, point-wise annotations, pair-wise comparisons, and list-wise comparisons. Common evaluation metrics include accuracy (for tasks like mathematical reasoning found in GSM8K and MATH), win rates derived from pairwise comparisons (such as MT-Bench and Anthropic-HH), and attack success rates used for assessing adversarial robustness (AdvBench).

\subsection{Results}
To demonstrate the effectiveness of different DPO variants, we conduct experiments on the Open LLM Leaderboard.
We compare different DPO variants using Mistral-7B-Base, Mistral-7B-Instruct~\cite{jiang2023identifying}, LLaMA-3-8B-Base, and LLaMA-3-8B-Instruct~\cite{dubey2024llama} as starting points.
The overall experimental setup follows \citet{Meng2024simpo}, ensuring a reproducible evaluation of different DPO variants.
For Mistral-7B-Base and LLaMA-3-8B-Base, the SFT models are trained based on the UltraChat-200k dataset~\cite{ding2023enhancing}, and subsequently applied different DPO variants on the SFT models using the UltraFeedback dataset~\cite{cui2023ultrafeedback}.
For Mistral-7B-Instruct and LLaMA-3-8B-Instruct, which have already undergone instruction-tuning, the preference dataset is regenerated by collecting responses from the SFT models using prompts from the UltraFeedback dataset~\cite{cui2023ultrafeedback}.

The experimental results, as summarized in Table~\ref{tab:results1}, highlight the performance of different DPO variants across various benchmarks. For the Mistral-7B-Base and LLaMA-3-8B-Base models, ORPO consistently achieves the highest average scores, indicating its effectiveness in aligning models with human preferences. Notably, ORPO outperforms other methods on IFEval, BBH, and MATH, demonstrating its superiority in instruction-following and mathematical reasoning tasks. Meanwhile, SLiC-HF and KTO also achieve competitive results, particularly in BBH and GPQA, suggesting that these methods effectively leverage preference data for enhanced performance. For the Mistral-7B-Instruct and LLaMA-3-8B-Instruct models, the improvements across different DPO variants are more nuanced. While DPO and R-DPO show strong performance in IFEval and MMLU-Pro, IPO and CPO demonstrate robustness in handling complex reasoning tasks like MATH and GPQA. Overall, the results indicate that different DPO variants exhibit varying strengths across benchmarks, with some methods excelling in base models while others are more effective for instruct models.

\begin{table*}[!t]
    \centering
    \caption{Experimental results of different DPO variants on Open LLM Leaderboard. The underline indicates the best performance.}
    \vspace{-0.2cm}
    \resizebox{\textwidth}{!}{
    \begin{tabular}{lcccccccccccccc}
        \toprule
        \multirow{2}{*}{\textbf{Model}} & \multicolumn{7}{c}{\textbf{Mistral-7B-Base}}  & \multicolumn{7}{c}{\textbf{LLaMA-3-8B-Base}} \\

        \cmidrule(lr){2-8} \cmidrule(lr){9-15} 
        &\textbf{IFEval} & \textbf{BBH} & \textbf{MATH} & \textbf{GPQA} & \textbf{MUSR} & \textbf{MMLU-Pro} & \textbf{AVERAGE}
        &\textbf{IFEval} & \textbf{BBH} & \textbf{MATH} & \textbf{GPQA} & \textbf{MUSR} & \textbf{MMLU-Pro} & \textbf{AVERAGE} \\
        
        \midrule
        SFT &3.4 &41.1 &9.2 &28.8 &42.0 &27.7   &25.4                &29.0 &46.3 &15.3 &28.6 &41.3 &31.0 &31.9 \\
        RRHF~\cite{yuan2023rrhf} &10.0 &40.6 &1.7 &26.4 &\ul{46.3} &26.1  &25.2               &31.0 &46.8 &13.9 &31.4 &36.8 &30.5 &31.7 \\
        SLiC-HF~\citep{zhao2023slic} &11.0 &44.0 &9.9 &29.2 &42.6 &28.1 &27.5             &\ul{41.7} &\ul{49.5} &17.5 &30.4 &39.7 &31.7 &\ul{35.1} \\
        DPO~\cite{rafailov2024direct} &11.1 &43.7 &7.1 &28.5 &43.8 &26.7     &26.8             &34.3 &48.2 &17.2 &31.9 &40.1 &31.5 &33.9 \\
        IPO~\citep{azar2024general} &9.4 &42.8 &9.7 &29.7 &39.7 &27.8      &26.5             &35.3 &49.0 &15.9 &\ul{32.8} &\ul{41.4} &31.9 &34.4 \\
        CPO~\cite{xu2024contrastive} &8.0 &42.7 &9.6 &28.9 &42.1 &27.3     &26.4              &32.4 &46.9 &16.8 &30.6 &39.1 &31.8 &32.9 \\
        KTO~\cite{ethayarajh2024kto} &12.9 &43.7 &12.0 &28.9 &46.1 &28.3   &28.6              &40.2 &48.3 &\ul{18.0} &31.0 &40.1 &31.1 &34.8 \\
        ORPO~\cite{hong2024orpo} &\ul{28.4}&\ul{46.4} &\ul{13.5} &\ul{30.2} &41.4 &\ul{29.5}  &\ul{31.6}             &40.0 &49.1 &16.8 &30.7 &38.4 &\ul{32.0} &34.5 \\
        R-DPO~\cite{park2024disentangling} &10.0 &43.0 &7.6 &28.7 &39.3 &27.2  &26.0              &36.4 &48.8 &17.2 &31.6 &40.6 &31.5 &34.4 \\
        SimPO~\cite{Meng2024simpo} &11.1 &43.1 &8.4 &28.9 &39.5 &27.2  &26.4              &40.8 &48.6 &15.8 &31.0 &40.5 &31.8 &34.7 \\
        \midrule
        \multirow{2}{*}{\textbf{Model}} & \multicolumn{7}{c}{\textbf{Mistral-7B-Instruct}}  & \multicolumn{7}{c}{\textbf{LLaMA-3-8B-Instruct}} \\

        \cmidrule(lr){2-8} \cmidrule(lr){9-15} 
        &\textbf{IFEval} & \textbf{BBH} & \textbf{MATH} & \textbf{GPQA} & \textbf{MUSR} & \textbf{MMLU-Pro} & \textbf{AVERAGE}
        &\textbf{IFEval} & \textbf{BBH} & \textbf{MATH} & \textbf{GPQA} & \textbf{MUSR} & \textbf{MMLU-Pro} & \textbf{AVERAGE} \\
        
        \midrule
        SFT &48.4 &\ul{46.2} &10.9 &\ul{29.1} &47.6 &27.1   &\ul{34.9}            &50.7 &49.3 &26.9 &31.0 &37.9 &35.7 &38.6 \\
        RRHF~\cite{yuan2023rrhf} &45.2 &45.3 &10.1 &28.5 &44.2 &26.2   &33.3                    &\ul{51.3} &49.3 &\ul{27.2} &29.6 &\ul{39.5} &35.3 &\ul{38.7} \\
        SLiC-HF~\citep{zhao2023slic} &39.4 &\ul{46.2} &11.4 &28.7 &49.0 &26.8  &33.6         &41.6 &\ul{50.9} &26.3 &\ul{31.3} &39.2 &35.3 &37.4 \\
        DPO~\cite{rafailov2024direct} &\ul{49.0} &45.6 &11.0 &26.9 &46.1 &26.8   &34.2            &48.9 &50.1 &25.8 &29.4 &38.7 &\ul{36.0} &38.2 \\
        IPO~\citep{azar2024general} &42.6 &45.3 &\ul{11.8} &27.8 &\ul{49.3} &27.2   &34.0            &50.4 &49.5 &26.3 &29.6 &37.9 &35.7 &38.2 \\
        CPO~\cite{xu2024contrastive} &38.8 &46.0 &10.1 &28.5 &48.4 &26.9   &33.1            &50.6 &49.1 &26.8 &\ul{31.3} &38.1 &35.8 &38.6 \\
        KTO~\cite{ethayarajh2024kto} &46.2 &45.7 &10.9 &27.8 &46.0 &27.3   &34.0            &43.1 &50.1 &26.3 &31.2 &38.1 &35.0 &37.3 \\
        ORPO~\cite{hong2024orpo} &37.6 &45.1 &11.2 &28.2 &46.9 &26.5   &32.6           &43.0 &50.6 &26.9 &29.3 &39.1 &35.1 &37.3 \\
        R-DPO~\cite{park2024disentangling} &46.8 &45.9 &9.9 &28.7 &46.2 &\ul{27.6}   &34.2           &50.9 &50.3 &25.3 &29.8 &39.0 &35.7 &38.5 \\
        SimPO~\cite{Meng2024simpo} &45.4 &45.9 &10.4 &28.3 &45.0 &27.1   &33.7          &48.8 &49.2 &25.0 &29.3 &39.2 &35.1 &37.8 \\
        \bottomrule
    \end{tabular}
    }
    \vspace{-0.3cm}
    \label{tab:results1}
\end{table*}

\section{Applications}\label{sec:applications}

In this section, we discuss the applications of DPO in various domains, including different LLM-based applications, diffusion models, and multi-modal LLMs.
We provide an overview of the key challenges and opportunities in each domain and highlight the potential impact of DPO on real-world applications.

\subsection{LLM-based Applications}

DPO has emerged as a powerful paradigm for aligning LLMs with human preferences across diverse applications~\cite{xu2024contrastive,wu2024word,hu2025fine,dubey2024llama}. In code generation, DPO enhances control over code quality by optimizing based on preferences from automated tests~\cite{gee2024code,miao2024aligning,zhang2024codedpo}. In mathematical reasoning, DPO reduces errors in complex problem-solving by emphasizing step-level preference optimization~\cite{chen2024step,wang2024self,lai2024step,lu2024step}. Multilingual applications leverage DPO to synchronize cross-lingual preferences, thereby improving translation accuracy and cultural relevance~\cite{she2024mapo,lai2024llms}. Recommendation systems utilize DPO to refine personalization by incorporating user preference data to optimize item rankings, thereby enhancing the model ability to distinguish preferred items from less preferred ones~\cite{chen2024softmax,bai2024finetuning}. These examples highlight the adaptability of DPO in achieving human-aligned outputs across diverse tasks.

\subsection{Diffusion Models}

In the realm of diffusion models, DPO has been adapted to better align generated content with user expectations~\cite{gu2024diffusion,shekhar2024see,li2025aligning,majumder2024tango}. By optimizing preferences over image-text pairs, DPO enhances the semantic accuracy of generated images and mitigates the production of undesirable or biased content. Studies have demonstrated that diffusion models fine-tuned with DPO respond more accurately to complex prompts compared to those trained with traditional techniques. Moreover, the efficiency of DPO allows for the fine-tuning of large-scale models using limited preference data, addressing significant computational challenges in training diffusion models~\cite{wallace2024diffusion,yang2024a,yang2024using}. While scaling DPO for high-resolution and dynamic content generation remains challenging, its ability to simplify reward modeling makes it a promising method for controlled content creation~\cite{liu2024alignment}.

\subsection{Multi-Modal LLMs}

For multi-modal LLMs, DPO plays a crucial role in aligning preferences across different data types, thereby improving coherence in tasks such as visual question answering and image captioning~\cite{li2024multi,wang2024mdpo,liang2024aligncap,amirloo2024understanding,fu2025chip}. By optimizing alignment between textual responses and visual inputs, DPO reduces hallucinations in multi-modal interactions, ensuring outputs remain faithful to the given context.
Although reconciling different types of feedback can be challenging, DPO offers a practical framework for lightweight adaptation, making it well-suited to preference-intensive multi-modal applications~\cite{zhang2024direct,xie2024v,li2024multi}.

\section{Challenges and Future Directions}\label{sec:challenges}

In this section, we discuss the key challenges and future directions in DPO research. We identify several critical issues that need to be addressed to further advance the field. Moreover, we propose several promising research directions that can help overcome these challenges and accelerate the adoption of DPO in the future.

\subsection{Efficient Preference Optimization}

Efficient preference optimization remains a pivotal challenge, as current DPO methods hinge on the availability of high-quality preference data, yet the manual collection of human annotations is both time-consuming and labor-intensive while automatically model-generated datasets often suffer from issues such as limited diversity, inherent biases, and insufficient fidelity to human judgment~\cite{guo2024direct,qi2024online,chen2024optune,wang2024self}. Moreover, even though DPO circumvents the intricacies of reward model engineering common in RL, it does not fully leverage the exploratory strengths that RL methods offer, as evidenced by recent advances in reasoning approaches where RL-based training has achieved notable successes~\cite{jaech2024openai,guo2025deepseek}. This opens up an avenue for future research to not only enhance data efficiency through advanced learning techniques but also to integrate novel exploration mechanisms~\cite{xie2025exploratory,bai2025online}, potentially through hybrid models that amalgamate the direct preference optimization benefits of DPO with the robust exploratory capabilities characteristic of RL.

\subsection{Multi-Modal Preference Optimization}

Multi-Modal Preference Optimization presents another frontier, given that existing DPO frameworks have primarily targeted text-based modalities while many real-world applications demand the alignment of diverse human preferences across text, images, audio, and even video~\cite{zhang2024direct,xie2024v,li2024multi,xu2024lvlm,huang2025causality}. In scenarios where cross-modal cues might conflict, such as the need for concise text paired with richly detailed imagery, the challenge lies in constructing a unified preference representation space that can intelligently and automatically recalibrate the priority of different modalities based on the contextual demands of the task at hand~\cite{wang2024mdpo,amirloo2024understanding,fu2025chip}. Future directions in this area could involve the development of innovative multi-modal preference encoding architectures, which are capable of disentangling compound preferences into modality-specific and cross-modal components that align conflicting preferences while also adapting dynamically to changing inputs.

\subsection{Continuous Preference Optimization}

Continuous preference optimization addresses the dynamic nature of human preferences that evolve over time or vary with different phases of a task, a factor that static DPO models often fail to capture~\cite{yuan2024selfrewarding,wang2024cream,liu2024direct,son2024right}. As social norms and individual preferences shift, there is an increasing need for systems that can continuously recalibrate their alignment strategies in real time while simultaneously mitigating the risk of catastrophic forgetting. Future research in this domain may focus on meta-learning approaches that enable models to learn not only from the current state of preferences but also how to efficiently adapt when these preferences change. By integrating online learning frameworks with mechanisms for detecting temporal shifts and contextual variability in user behavior, researchers can pave the way toward systems that remain consistently relevant and effective in the face of evolving societal and individual expectations.

\subsection{Interpretable Preference Optimization}

Interpretable preference optimization is critical for building trust in models that implicitly align human values, as the opaque nature of current DPO complicates the ability to audit and control the alignment process. In practice, human preferences are multi-dimensional~\cite{zhou2024beyond,lou2024spo,guo2024controllable}, encompassing aspects such as factual accuracy, fairness, creativity, and beyond, and there is a pressing need to decompose these complex preferences into interpretable components that can be individually examined and fine-tuned. Future research could leverage advances in explainable techniques to develop models that not only achieve fine-grained alignment across diverse values but also provide transparent insights into how different preference dimensions interact to shape final decisions. This level of interpretability would allow stakeholders to balance competing values more effectively, ensuring that the alignment process remains both accountable and adaptable as societal norms continue to evolve.

\section{Conclusion}\label{sec:conclusion}

In recent years, DPO has emerged as a promising paradigm for aligning LLMs with human preferences by directly optimizing model policies using preference data. Despite its potential, the DPO research landscape remains fragmented, with a lack of systematic organization and comparative analysis. In this survey, we present a comprehensive overview of DPO and introduce a novel taxonomy that categorizes existing works into four key dimensions: data strategy, learning framework, constraint mechanism, and model property. We have also discussed the key benchmarks, evaluation results, and applications of DPO, highlighting the challenges and future directions in this field. By providing a systematic analysis of the existing DPO methods, we aim to facilitate further research and development in this area.

\footnotesize
\bibliographystyle{myunsrtnat}
\bibliography{ref}

\vfill

\end{document}

%% file: fig/overview.tex
\definecolor{mc1}{RGB}{109,173,209}
\definecolor{mc2}{RGB}{182,215,232}
\definecolor{mc3}{RGB}{233,241,244}
\definecolor{mc4}{RGB}{233,241,244}
\definecolor{mcleaf}{RGB}{251,227,213}
\definecolor{data}{HTML}{F8CECC}
\definecolor{learn}{HTML}{DAE8FC}
\definecolor{constraint}{HTML}{D5E8D4}
\definecolor{model}{HTML}{FFF2CC}
\definecolor{dpo}{HTML}{F9F7ED}

\tikzstyle{my-box}=[
    rectangle,
    rounded corners,
    text opacity=1,
    minimum height=1.5em,
    minimum width=6em,
    inner sep=3pt,
    align=center,
    fill opacity=.3,
    thick,
]
\tikzstyle{leaf}=[
    my-box, 
    fill=gray!3, 
    text=black, 
    align=left,
    text width=53em,
    font=\normalsize,
    inner xsep=3pt,
    inner ysep=4pt,
]
\begin{figure*}[!t]
    \centering
    \resizebox{\textwidth}{!}{
        \begin{forest}
            forked edges,
            for tree={
                grow=east,
                reversed=true,
                anchor=base west,
                parent anchor=east,
                child anchor=west,
                base=center,
                font=\normalsize,
                rectangle,
                draw=gray,
                rounded corners,
                align=left,
                text centered,
                minimum width=5em,
                edge+={gray!60, thick},
                s sep=4pt,
                inner xsep=3pt,
                inner ysep=3pt,
                thick,
                fit=band,
            },
            where level=0{
                folder,
                grow'=0,
                fill=dpo,
                font=\large,
                child anchor=west,
                parent anchor=south west,
                anchor=west,
                calign=first,
                yshift=120pt,
            }{},
            where level=1{
                text width=14em,
            }{},
            where level=2{text width=10em}{},
            where level=3{text width=9em}{},
            [
                {\quad\quad\qquad\qquad\quad{\textbf{A Taxonomy of Direct Preference Optimization (DPO)}}\qquad\qquad\qquad
                \begin{minipage}[h]{0.68\textwidth}
                    \centering
                    \includegraphics[width=\linewidth]{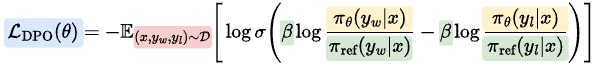}
                \end{minipage}\qquad\qquad\qquad\qquad}
                [
                    \textbf{Data Strategy (\S \ref{sec:data})}, fill=data
                    [
                        \textbf{Quality (\S \ref{sec:quality})}, fill=data
                        [
                            \textbf{Heterogeneity}, fill=data
                            [   
                                MinMax-DPO~\cite{chidambaram2024direct}{, }
                                MallowsPO~\cite{chen2024mallowspo}{, }
                                GRPO~\cite{ramesh2024group}{, }
                                GDPO~\cite{yao2025no}
                                , leaf
                            ]
                        ]
                        [
                            \textbf{Distinguishability}, fill=data
                            [
                                ODPO~\cite{amini2024direct}{,} 
                                MPO~\cite{gou2024mixed}{,} 
                                DPO-rc~\cite{wang2024reward}{,} 
                                $\alpha$-DPO~\cite{wu2024alpha}{,}
                                GDPO~\cite{furuta2024geometric}{,} 
                                $\beta$-DPO~\cite{wu2024betadpo}{,}
                                fDPO~\cite{morimura2024filtered}{, } 
                                Curri-DPO~\cite{pattnaik2024enhancing}{,} \\
                                Ada-DPO~\cite{hong2024adaptive}{,} 
                                sDPO~\cite{kim2024sdpo}{,} 
                                Sr-DPO~\cite{yu2024direct}{,} 
                                GaPO~\cite{jieming2024gapaware}{, }
                                \citet{ma2024plugandplay} 
                                , leaf
                            ]
                        ]
                        [
                            \textbf{Noise}, fill=data
                            [
                                rDPO~\cite{chowdhury2024provably}{,} 
                                PerpCorrect~\cite{kong2024perplexityaware}{,} 
                                ROPO~\cite{liang2024ropo}{,} 
                                SPA~\cite{kim2025spread}{,} 
                                \citet{zhang2025combating}{,} 
                                \citet{im2025understanding}{,} 
                                \citet{gao2024impact}{,} 
                                \citet{wu2024towards} 
                                , leaf
                            ]
                        ]
                    ]
                    [
                        \textbf{Feedback  (\S \ref{sec:feedback})}, fill=data
                        [
                            \textbf{Point-Wise}, fill=data
                            [                            
                                KTO~\cite{ethayarajh2024kto}{, }
                                BCO~\cite{jung2024binary}{, }
                                Cal-DPO~\cite{globerson2024cal}{, }
                                DRO~\cite{wu2024towards}{, }
                                AOT~\cite{melnyk2024distributional}{, }
                                ULMA~\cite{cai2023ulma}{, }
                                NCA~\cite{chen2024noise}{, } 
                                GPO~\cite{zhang2024general}
                                , leaf
                            ]
                        ]
                        [
                            \textbf{Pair-Wise}, fill=data
                            [
                                DPO~\cite{rafailov2024direct}{,}
                                IPO~\cite{azar2024general}{,}
                                DPO-RK~\cite{chen2024extending}{,} 
                                DPO-D~\cite{chen2024extending}{,}
                                BMC~\cite{jiang2024bridging}{,}
                                IOPO~\cite{zhang2024iopo}{,}
                                PMPO~\cite{abdolmaleki2024preference}{,} 
                                SAPO~\cite{yin2024self}{, }
                                D2O~\cite{duan2024negating}{, }\\
                                NPO~\cite{zhang2024negative}{, }
                                SimNPO~\cite{fan2024simplicity}
                                , leaf
                            ]
                        ]
                        [
                            \textbf{List-Wise}, fill=data
                            [
                                Panacea~\cite{zhong2024panacea}{,}
                                LiPO~\cite{liu2402lipo}{,}
                                LIRE~\cite{zhu2024lire}{,}
                                OPO~\cite{zhao2024ordinal}{,}
                                DRPO~\cite{zhou2024optimizing}{,}
                                mDPO~\cite{wang2024mdpo}{,} 
                                RPO~\cite{yin2024relative}{,} 
                                TODO~\cite{guo2024todo}
                                , leaf
                            ]
                        ]
                    ]
                    [
                        \textbf{Granularity  (\S \ref{sec:granularity})}, fill=data
                        [
                            \textbf{Token-Level}, fill=data
                            [    
                                \citet{rafailov2024from}{, }
                                TDPO~\cite{zeng2024token}
                                TIS-DPO~\cite{liu2024tis}{, }
                                SparsePO~\cite{christopoulou2024sparsepo}{, }
                                RTO~\cite{zhong2024dpo}{, }
                                SePO~\cite{yang2024selective}{, } 
                                EPO~\cite{qi2024epo}{, }
                                D$^2$PO~\cite{shao2025earlier}
                                , leaf
                            ]
                        ]
                        [
                            \textbf{Step-Level}, fill=data
                            [
                                Step-DPO~\cite{lai2024step}{,} 
                                SCDPO~\cite{lu2024step}{,} 
                                CPO~\cite{zhang2024chain}{,} 
                                MCTS-DPO~\cite{xie2024monte}{,} 
                                TPO~\cite{liao2024tpo}{,} 
                                RDPO~\cite{just2024data}{,}  
                                DAPO~\cite{liu2024improving}
                                , leaf
                            ]
                        ]
                        [
                            \textbf{Sentence-Level}, fill=data
                            [
                                DPO~\cite{rafailov2024direct}{,} 
                                MAPO~\cite{she2024mapo}{,} 
                                EURUS~\cite{yuan2024advancing}{,} 
                                IRPO~\cite{pang2025iterative}{,} 
                                FACTALIGN~\cite{huang2024factalign}
                                , leaf
                            ]
                        ]
                        [
                            \textbf{Turn-Level}, fill=data
                            [
                                M-DPO~\cite{xiong2024building}{,} 
                                ETO~\cite{song2024trial}{,} 
                                SDPO~\cite{kong2025sdpo}{,} 
                                AgentQ~\cite{putta2024agent}{, }
                                DMPO~\cite{shi2024direct}
                                , leaf
                            ]
                        ]
                    ]
                ]
                [
                    \textbf{Learning Framework (\S \ref{sec:learning})}, fill=learn
                    [
                        \textbf{Paradigm  (\S \ref{sec:paradigm})}, fill=learn
                        [
                            \textbf{Offline}, fill=learn
                            [                            
                                DPO~\cite{rafailov2024direct}{,}
                                CPO~\cite{xu2024contrastive}{,}
                                ORPO~\cite{hong2024orpo}{,}
                                ULMA~\cite{cai2023ulma}{,}
                                PAFT~\cite{pentyala2024paft}{,} 
                                Curri-DPO~\cite{pattnaik2024enhancing}{,} 
                                sDPO~\cite{kim2024sdpo}{,} \\
                                Linear Alignment~\cite{gao2024linear}{,}
                                ICDPO~\cite{song2024icdpo}
                                , leaf
                            ]
                        ]
                        [
                            \textbf{Online}, fill=learn
                            [
                                OAIF~\cite{guo2024direct}{,}
                                OFS-DPO~\cite{qi2024online}{,}
                                \citet{yuan2024selfrewarding}{,}
                                BPO~\cite{xu2024bpo}{,}
                                RS-DPO~\cite{khaki2024rs}{,} 
                                RSO~\cite{liu2024statistical}{,} 
                                ROPO~\cite{liang2024ropo}{,}
                                \citet{shi2024crucial}{,} \\
                                OPTUNE~\cite{chen2024optune}{,}
                                Iterative RPO~\cite{pang2025iterative}{,} 
                                DPO-ST~\cite{wang2024self}{,}
                                \citet{xie2024monte}{,} 
                                APO~\cite{he2024accelerated}{,}
                                \citet{xiong2023iterative}{,}
                                COMAL~\cite{liu2024comal}{,}\\
                                \citet{xu2024things}{,}
                                SeRA~\cite{ko2024sera}{,}
                                CREAM~\cite{wang2024cream}{,}
                                D2PO~\cite{singhal2024dpo}{,}
                                DLMA~\cite{liu2024direct}{,}
                                XPO~\cite{xie2025exploratory}{,}
                                SELM~\cite{zhang2024self}{,}
                                ETO~\cite{song2024trial}{,}\\
                                VPO~\cite{cen2024value}{,}
                                \citet{xiong2024building}{,}
                                COPO~\cite{bai2025online}{,}
                                HyPO~\cite{song2024importance}{,}
                                MPO~\cite{gou2024mixed}{,}
                                AIPO~\cite{shen2024aipo}{,}
                                \citet{tang2024understanding}{,} 
                                \citet{xu2024is}
                                , leaf
                            ]
                        ]
                        [
                            \textbf{Active}, fill=learn
                            [
                                APL~\cite{muldrew2024active}{,}
                                divAPO~\cite{choi2024active}{,}
                                ADPO\cite{ji2024reinforcement}{,}
                                \citet{das2024active}
                                , leaf
                            ]
                        ]
                    ]
                    [
                        \textbf{Objective  (\S \ref{sec:objective})}, fill=learn
                        [
                            \textbf{Multi-Objective}, fill=learn
                            [
                                MODPO~\cite{zhou2024beyond}{,}
                                SPO~\cite{lou2024spo}{,}
                                MOSLIM~\cite{zhang2025moslimalign}{,}
                                HPO~\cite{badrinath2024hybrid}{,}
                                CPO~\cite{guo2024controllable}{,}
                                DRDO~\cite{nath2024simultaneous}
                                , leaf
                            ]
                        ]
                        [
                            \textbf{Self-Play}, fill=learn
                            [                            
                                SPIN~\cite{chen2024self}{,} 
                                SPPO~\cite{wu2024self} {,} 
                                SPO~\cite{swamy2024minimaximalist}{,} 
                                BoNBoN~\cite{gui2406bonbon}{,} 
                                Nash-MD~\cite{munos2023nash}{,}  
                                DNO~\cite{rosset2024direct}{,}
                                IPO-MD~\cite{calandriello2024human}{,} \\
                                SRPO~\cite{choi2024self}{,}
                                DNPO~\cite{yang2025dynamic}
                                , leaf
                            ]
                        ]
                    ]
                ]
                [
                    \textbf{Constraint Mechanism (\S \ref{sec:constraint})}, fill=constraint
                    [
                        \textbf{Reference  (\S \ref{sec:reference})}, fill=constraint
                        [
                            \textbf{Dynamic}, fill=constraint
                            [                            
                                Curri-DPO~\citep{pattnaik2024enhancing}{,} 
                                MPO~\citep{gou2024mixed}{,} 
                                M-DPO~\citep{wang2024mdpo}{,} 
                                \citet{gorbatovski2024learn}
                                , leaf
                            ]
                        ]
                        [
                            \textbf{Free}, fill=constraint
                            [
                                ORPO~\cite{hong2024orpo}{, } 
                                SimPO~\cite{Meng2024simpo}{, }
                                SimPER~\cite{xiao2025simper}{, }
                                \citet{liu2024understanding}{, }
                                \citet{xu2024contrastive}
                                , leaf
                            ]
                        ]
                    ]
                    [
                        \textbf{Divergence  (\S \ref{sec:divergence})}, fill=constraint
                        [
                            \textbf{Diversity}, fill=constraint
                            [                            
                                f-DPO~\cite{wang2023beyond}{, } \citet{slocum2025diverse}
                                , leaf
                            ]
                        ]
                        [
                            \textbf{Generalization}, fill=constraint
                            [
                                DPO-Kernels~\cite{das2025dpo}{,}
                                FlipGuard~\cite{zhu2024flipguard}{,}
                                FPO~\cite{yin2024direct}{,}
                                GPO~\cite{tang2024generalized}{,} 
                                EXO~\cite{ji2024towards}{, } 
                                SPO~\cite{sharifnassab2024soft}{, }
                                QDPO~\cite{lee2024improving}{, }
                                \citet{huang2025correcting}
                                , leaf
                            ]
                        ]
                    ]
                    [
                        \textbf{Safety  (\S \ref{sec:safety})}, fill=constraint
                        [
                            SafeDPO~\cite{kim2025safedpo}{, }
                            SACPO~\cite{akifumi2024stepwise}{, }
                            C-DPO~\cite{liu2024enhancing}{, }
                            ADPO~\cite{kim2024adversarial}{, }
                            \citet{lee2024amechanistic}{, }
                            \citet{zhang2025backtracking}
                            , leaf, text width=64em
                        ]
                    ]
                ]
                [
                    \textbf{Model Property (\S \ref{sec:model})}, fill=model
                    [
                        \textbf{Generation  (\S \ref{sec:generation})}, fill=model
                        [
                            \textbf{Distribution}, fill=model
                            [
                                NS-DPO~\cite{son2024right}{,} 
                                SRPO~\cite{choi2025selfimproving}{,}
                                SELM~\cite{rafailov2024direct}{,}
                                e-DPO~\cite{fisch2024robust}{,}
                                \citet{lin2024limited}{,}
                                \citet{tajwar2024preference}
                                , leaf
                            ]
                        ]
                        [
                            \textbf{Length}, fill=model
                            [
                                RRHF~\cite{yuan2023rrhf}{,} 
                                R-DPO~\cite{park2024disentangling}{,} 
                                SimPO~\cite{Meng2024simpo}{,} 
                                SamPO~\cite{lu2024eliminating}{,} 
                                DMPO~\cite{shi2024direct}{,} 
                                LIFT~\cite{yuan2024following}{,} 
                                LN-DPO~\cite{ahrabian2024hitchhiker}{,} 
                                LD-DPO~\cite{liu2024length}{,} \\
                                D$^2$PO~\cite{shao2025earlier}{,} 
                                LongPO~\cite{chen2025longpo}{,} 
                                \citet{singhal2024a}{,} 
                                \citet{richardson2024declarative}
                                , leaf
                            ]
                        ]
                    ]
                    [
                        \textbf{Optimization  (\S \ref{sec:optimization})}, fill=model
                        [
                            \textbf{Likelihood}, fill=model
                            [                            
                                APO~\cite{doosterlinck2024anchored}{,}
                                DPO-Positive~\cite{pal2024smaug}
                                e-DPO~\cite{fisch2024robust}{,} 
                                Flex-DPO~\cite{yan2025dproperties}{,} 
                                \citet{rafailov2024from}{,}
                                \citet{feng2024towards}{,}
                                \citet{yuan2025common}{,} \\
                                \citet{razin2024unintentional}{,}
                                \citet{shi2024understanding}
                                , leaf
                            ]
                        ]
                        [
                            \textbf{Alignment}, fill=model
                            [
                                PAFT~\cite{pentyala2024paft}{,} 
                                \citet{lin2024mitigating}{,} 
                                \citet{thakkar2024deep}{,} 
                                \citet{lu2024online}{,}
                                \citet{chen2024preference} 
                                , leaf
                            ]
                        ]
                    ]
                ]
            ]
        \end{forest}
    }
    \caption{A taxonomy of DPO. We categorize existing DPO works into four branches: \textit{data strategy}, \textit{learning framework}, \textit{constraint mechanism}, and \textit{model property}. Different colored boxes indicate different categories and their corresponding representative references.}
    \vspace{-0.4cm}
    \label{fig:overview}
\end{figure*}